\documentclass{article}
\pdfoutput=1  %

\usepackage[main, final]{neurips_2026}

\usepackage[utf8]{inputenc}
\usepackage[T1]{fontenc}
\usepackage{hyperref}
\usepackage{url}
\usepackage{booktabs}
\usepackage{amsfonts}
\usepackage{amsmath}
\usepackage{nicefrac}
\usepackage{microtype}
\usepackage{graphicx}
\usepackage{float}
\usepackage{placeins}
\usepackage{xspace}
\usepackage{xcolor}
\usepackage{tabularx}
\usepackage{xltabular}
\usepackage[breakable, skins]{tcolorbox}
\newtcolorbox{promptbox}[1][]{
  enhanced,
  breakable,
  colback=gray!4,
  colframe=gray!40,
  colbacktitle=gray!20,
  boxrule=0.5pt,
  arc=2pt,
  left=10pt, right=10pt, top=8pt, bottom=8pt,
  fonttitle=\bfseries\sffamily\small,
  coltitle=black,
  attach boxed title to top left={xshift=8pt, yshift=-8pt},
  boxed title style={
    colback=gray!20,
    colframe=gray!40,
    boxrule=0.5pt,
    arc=2pt,
    left=6pt, right=6pt, top=2pt, bottom=2pt,
  },
  title=#1,
  fontupper=\ttfamily\small,
}
\usepackage{epigraph}
\newcommand{\modelname}{\textbf{CORAL}\xspace}

\title{\modelname: Learning Amyloid Fibril Ligand Docking with Cooperative Binding Rewards}

\author{%
  Yasheng Sun$^{1*\dagger}$ \quad
  Bohan Li$^{2*}$ \quad
  Youqi Tao$^{3\dagger}$ \quad
  Jürgen Schmidhuber$^{1}$ \\[0.6em]
  $^{1}$Center of Excellence for Generative AI, KAUST \\[0.15em]
  $^{2}$Department of Biomedical Engineering, National University of Singapore \\[0.15em]
  $^{3}$Stern Laboratory, Brigham and Women's Hospital, Harvard University
}

\begin{document}

\maketitle

\begingroup
\renewcommand{\thefootnote}{\fnsymbol{footnote}}
\begin{NoHyper}
\footnotetext[1]{Equal contribution.}
\footnotetext[2]{Corresponding authors.}
\end{NoHyper}
\endgroup
\setcounter{footnote}{2}

\begin{figure}[h]
\centering
\includegraphics[width=\linewidth]{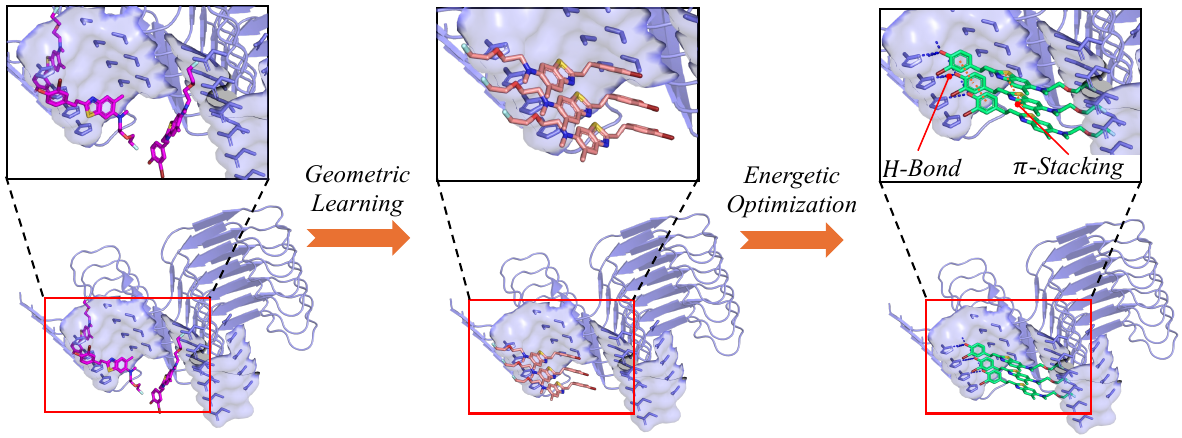}
\caption{\textbf{Overview of \modelname.} A two-stage training recipe progressively shapes the model's ligand-pose distribution on amyloid fibrils. The \emph{geometric learning} stage (left to middle) supervises the model on curated stacking poses, teaching it to align ligands along the cross-$\beta$ groove into periodic stacking arrangements. The \emph{energetic optimisation} stage (middle to right) further refines the pose distribution via online reinforcement learning under a composite reward that combines protein--ligand binding affinity with cooperative inter-ligand stacking energy, leading to tighter contacts such as the H-bond and $\pi$--$\pi$ stacking interactions highlighted in the right inset.}
\label{fig:teaser}
\end{figure}

\begin{abstract}
A hallmark of neurodegenerative diseases such as Alzheimer's and Parkinson's is the aberrant aggregation of proteins into amyloid fibrils, and small molecules that selectively bind to these fibrils hold promise as diagnostics, imaging probes, and therapeutics.
Predicting how such ligands bind to fibril targets, however, presents two fundamental challenges.
First, resolved co-crystal structures of amyloid--ligand complexes are exceptionally scarce; even with recent advances in cryo-EM only a handful have been structurally characterized, making supervised training of docking models impractical for this target class.
Second, amyloid fibrils present a binding mode fundamentally different from globular proteins: ligands intercalate into longitudinal cross-$\beta$ grooves and stack cooperatively along the fibril axis, a geometry that existing docking models are not designed to capture.
To address these challenges, we present \modelname (\textbf{CO}ope\textbf{R}ative \textbf{A}myloid \textbf{L}igand docking), a reinforcement learning framework that trains a generative docking model to produce ligand pose distributions tailored to the cross-$\beta$ groove geometry.
Our reward explicitly incorporates cooperative ligand--ligand stacking energy alongside protein--ligand docking affinity, directly capturing the distinctive binding geometry of amyloid fibrils.
We further introduce a curated evaluation set of amyloid--ligand complexes constructed from model-generated poses validated by domain experts.
Experiments on both experimentally resolved structures and this evaluation set demonstrate improved pose quality and binding affinity correlation over existing docking baselines.
Our project page is available at {\hypersetup{hidelinks}\href{https://sunyasheng.github.io/projects/CORAL}{\textcolor[HTML]{D12A8A}{sunyasheng.github.io/projects/CORAL}}}.
\end{abstract}

\section{Introduction}

\begin{flushright}
\begin{minipage}{0.85\linewidth}
\small\itshape ``Of his bones are coral made; nothing of him that doth fade, but doth suffer a sea-change into something rich and strange.''\hfill\normalfont\small ---~Shakespeare, \textit{The Tempest} (1611)
\end{minipage}
\end{flushright}
\vspace{0.5em}

Predicting how a small molecule binds to a target protein, known as the molecular docking problem, is a cornerstone of computational drug discovery, and recent deep learning approaches have achieved remarkable progress on globular protein targets~\cite{corso2023diffdock,qiao2024neuralplexer,abramson2024alphafold3}.
However, amyloid fibrils (the insoluble protein aggregates implicated in Alzheimer's, Parkinson's, and related neurodegenerative diseases) represent an increasingly important yet largely neglected class of docking targets.
Developing small molecules that selectively bind amyloid fibrils is critical for both diagnostics (PET tracers, imaging probes) and therapeutics~\cite{chisholm2024closer,chisholm2025ligands}, yet predicting how a given compound docks to a fibril remains an open challenge.

Modelling the binding modes of amyloid fibril--ligand complexes, however, faces two fundamental challenges.
\textit{First,} unlike globular proteins, amyloid fibrils cannot be crystallised, and cryo-EM, the only viable route to atomic-resolution fibril structures, often lacks sufficient resolution to unambiguously identify bound small-molecule ligands~\cite{ghosh2021cryoem_amyloid}; co-crystal structures of amyloid--ligand complexes therefore remain exceptionally rare.
Existing computational approaches~\cite{smith2024symdock,smith2025cooperative} rely on predefined potential energy functions to enumerate and score ligand poses; such physics-based methods are inherently limited by the brittleness of hand-crafted scoring functions and the rigid-body docking assumption.
\textit{Second,} unlike globular proteins, where a single ligand occupies a well-defined pocket, amyloid fibrils exhibit collective binding: multiple ligands intercalate cooperatively into cross-$\beta$ grooves, with inter-ligand $\pi$--$\pi$ stacking often dominating over direct protein contacts~\cite{merz2023stacked}.
The prevalence of fibril polymorphism, where the same protein can adopt multiple distinct fibril folds, further compounds this challenge, as the binding groove geometry varies across polymorphs.
While recent work~\cite{guo2025ribbonfold} has applied learning-based methods to model amyloid fibril structures, how to model cooperative multi-ligand binding within these structures remains unexplored.

To address these challenges, we propose \modelname (\textbf{CO}ope\textbf{R}ative \textbf{A}myloid \textbf{L}igand docking), a cooperative multi-ligand docking framework that \emph{mimics collaborative amyloid--ligand binding behaviours without demanding real-world co-crystal structures}.
Our key insight is to design a mechanism that encourages the model to explore how ligands interact with the fibril groove, so that generated structures naturally satisfy the cooperative binding mode.
Because co-crystal structures are scarce, we first construct a synthetic training set: diverse small-molecule conformers are generated with a drug generative model~\cite{schneuing2025drugflow}, each conformer is replicated along the fibril axis to approximate cooperative stacking, and rule-based filters are applied to curate plausible binding poses.
A structure generative model built on an AlphaFold\,3-style Pairformer--Diffusion architecture~\cite{abramson2024alphafold3} is supervised on this curated set, providing the model with a basic understanding of groove-intercalation geometry.
To further align the generated poses with the energetics of cooperative binding, we refine the model via online reinforcement learning on the diffusion process~\cite{diffusionnft2025}, guided by a composite reward that jointly considers protein--ligand binding affinity and inter-ligand energetic complementarity along the periodic fibril axis.
On the modelling side, amyloid fibrils are ribbon-like assemblies of repeated $\beta$-strand layers with a characteristic inter-chain C$\alpha$--C$\alpha$ spacing of ${\sim}4.8$\,\AA.
To faithfully capture this periodicity, we encode the inter-chain distance constraints directly into the Pairformer pair representation as an explicit structural prior for the fibril architecture.

Our main contributions can be summarised as follows:
\textbf{1)} We present \modelname, to our knowledge the first deep learning-based docking framework for amyloid fibril--ligand complexes.
\textbf{2)} We design a two-stage training recipe that enables the model to learn fibril-specific binding modes: a warm-up stage on synthetically curated stacking poses, followed by online reinforcement-learning fine-tuning with a cooperative binding reward.
\textbf{3)} We evaluate on two complementary datasets: a public set of fibril--ligand co-crystal structures and an expert-curated set whose reference poses are generated by computational agents and validated by domain experts. \modelname produces higher-quality poses than existing docking baselines on both.

\section{Related Work}

\paragraph{Amyloid Fibril Structures and Their Ligands.}
Small-molecule binding to tau, $\alpha$-synuclein, and amyloid-$\beta$ fibrils is central to diagnostics and therapeutics for neurodegenerative diseases~\cite{chisholm2024closer,chisholm2025ligands,polymorphism2023}.
Cryo-EM structures reveal that aromatic ligands intercalate into longitudinal fibril grooves and stack cooperatively, with ligand--ligand $\pi$--$\pi$ interactions often dominating over protein contacts~\cite{merz2023stacked,kunach2024mk6240,smith2024symdock,chisholm2024closer,chisholm2025ligands,asyn2024pnas}.
SymDOCK~\cite{smith2024symdock} extends classical docking to predict symmetric ligand stacks, reproducing experimentally determined tracer geometries, while a cooperative-binding model~\cite{smith2025cooperative} provides a thermodynamic framework for fibril--ligand affinities.
RibbonFold~\cite{guo2025ribbonfold} adapts AlphaFold2 to generate polymorph-diverse amyloid structures but models only a single protofilament, and benchmarks show that general-purpose co-folding models still struggle with fibril targets~\cite{af3amyloids2024}.
However, learning-based modelling of fibril--ligand binding modes remains an open problem.

\paragraph{Protein-Ligand Docking.}
Classical docking methods~\cite{jones1997gold,friesner2004glide,trott2010autodock,eberhardt2021autodock} score conformational samples with hand-crafted energy functions, yielding physically interpretable predictions~\cite{buttenschoen2024posebusters} but degrading in non-standard contexts~\cite{koes2013smina,mcnutt2021gnina}.
Learning-based methods replace these hand-crafted components with data-driven models: regression approaches~\cite{stark2022equibind,lu2022tankbind,zhang2023e3bind,zhou2023unimol,pei2023fabind} predict poses in a single forward pass, while diffusion-based docking~\cite{corso2023diffdock,corso2024diffdockl,lu2024dynamicbind,qiao2024neuralplexer,morehead2025flowdock} and co-folding methods~\cite{jumper2021alphafold2,abramson2024alphafold3,krishna2024rosettafoldaa,boitreaud2024chai1,wohlwend2024boltz1} further improve structural accuracy~\cite{morehead2025posebench}, building on a long line of neural protein structure prediction~\cite{pollastri2002sspro,hochreiter2007homology,golkov2016contact}.
All data-driven approaches share the same fundamental limitation, however: trained predominantly on globular proteins, they fail to generalise to non-standard binding geometries~\cite{morehead2025posebench,buttenschoen2024posebusters}, leaving amyloid fibril targets, with their distinctive cross-$\beta$ groove binding mode, largely out of reach.

\paragraph{Reinforcement Learning for Structure Generation.}
Conditional diffusion models have been applied to controllable image, motion, and video generation~\cite{sun2023imagebrush,sun2024avitalking,sun2025dreamconnect,sun2026coshdit}.
For molecular structures, early work adopted offline preference optimisation: AliDiff~\cite{gu2024alidiff}, DecompDPO~\cite{cheng2024decompdpo}, and AbDPO~\cite{zhou2024abdpo} extended DPO~\cite{rafailov2023dpo} to molecule, structure-based drug, and antibody design respectively.
The recent extension of GRPO~\cite{shao2024deepseekmath,deepseekr1,schmidhuber2015learningtothink,schmidhuber2018onebignet,schmidhuber1992history} to continuous diffusion and flow matching~\cite{xue2025dancegrpo,liu2025flowgrpo} has opened the door to \emph{online} RL fine-tuning of structural generative models, with concurrent work applying it to tasks such as protein inverse folding~\cite{wang2025proteinzero} and showing that online RL improves task-specific performance beyond what distribution matching alone can achieve.

\section{Methodology}

\subsection{Overview}

\paragraph{Problem Formulation.}
We formulate amyloid fibril--ligand docking as a co-folding problem~\cite{morehead2025posebench}.
An amyloid fibril is a periodic assembly of identical protein monomers (subunits) stacked along the fibril axis; we represent a fibril complex as $K_p$ copies of the monomer amino acid sequence $\mathbf{A}$ and $K_l$ copies of the same ligand molecular graph $\mathbf{G}$, where $K_p$ and $K_l$ may differ.
Given $(\mathbf{A}^{1:K_p}, \mathbf{G}^{1:K_l})$, we jointly sample the 3D heavy-atom coordinates of all protein monomers $\mathbf{x}_\text{prot} \in \mathbb{R}^{K_p N_p \times 3}$ and all ligands $\mathbf{x}_\text{lig} \in \mathbb{R}^{K_l N_l \times 3}$ from a generative model:
\begin{equation}
  (\mathbf{x}_\text{prot},\, \mathbf{x}_\text{lig}) \sim p_\theta(\mathbf{x}_\text{prot}, \mathbf{x}_\text{lig} \mid \mathbf{A}^{1:K_p}, \mathbf{G}^{1:K_l}),
\end{equation}
where $N_p$ and $N_l$ denote the number of heavy atoms per monomer and per ligand, respectively.

\paragraph{Preliminaries.}
\label{sec:preliminaries}
Recent structure prediction models~\cite{abramson2024alphafold3,wohlwend2024boltz1} adopt flow matching as their generative backbone due to its stable training and efficient sampling.
\textit{Flow matching}~\cite{lipman2023flow,liu2023flow} learns a velocity field $v_\theta(\mathbf{x}_t, t)$ that transports samples along a straight-line interpolant $\mathbf{x}_t = (1 - t)\,\mathbf{x}_0 + t\,\mathbf{x}_1$ between data $\mathbf{x}_0$ and noise $\mathbf{x}_1 \sim \mathcal{N}(\mathbf{0}, \mathbf{I})$, trained via:
\begin{equation}
  \mathcal{L}_\text{FM}(\theta) = \mathbb{E}_{t,\mathbf{x}_0,\mathbf{x}_1}\bigl[\| v_\theta(\mathbf{x}_t, t) - (\mathbf{x}_1 - \mathbf{x}_0) \|^2\bigr].
  \label{eq:flow_matching}
\end{equation}

\textit{DiffusionNFT}~\cite{diffusionnft2025} aligns a flow-matching model with downstream objectives by operating on the forward-process velocity field, avoiding sampler coupling and trajectory storage.
Given a scalar reward $r(\mathbf{x}_0)$, we obtain a weight $\tilde r(\mathbf{x}_0)\in[0,1]$ via group-relative scaling. Two implicit velocity targets are constructed from the current policy $v_\theta$ and the sampling policy $v_\text{old}$:
\begin{align}
  v^+_\theta &= (1 - \beta)\,v_\text{old} + \beta\,v_\theta, \\
  v^-_\theta &= (1 + \beta)\,v_\text{old} - \beta\,v_\theta.
\end{align}
The training objective is:
\begin{equation}
  \mathcal{L}_\text{NFT}(\theta) = \mathbb{E}\!\left[
    \tilde r \cdot \| v^+_\theta - v \|^2
    + (1 - \tilde r) \cdot \| v^-_\theta - v \|^2
  \right],
  \label{eq:nft_loss}
\end{equation}
where $v$ is the target velocity derived from the forward process.
High-reward samples are reinforced through $v^+_\theta$, while low-reward samples are suppressed through $v^-_\theta$.

\begin{figure}[t]
\centering
\includegraphics[width=\linewidth]{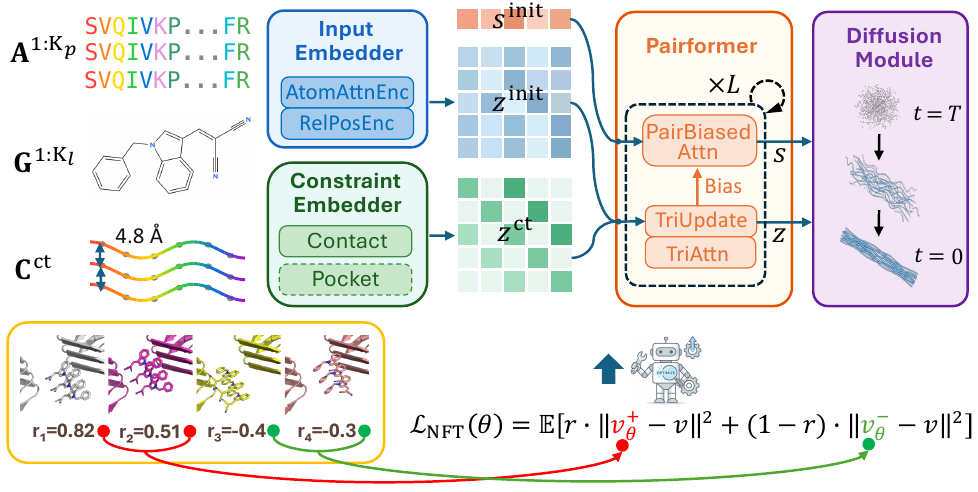}
\caption{Overview of \modelname. A Pairformer--Diffusion structure model with fibril-periodicity priors generates 3D fibril--ligand complexes from protein sequences and ligand graphs. Generated poses are scored by a reward that combines protein--ligand binding affinity with cooperative inter-ligand stacking energy, and the model is refined via online reinforcement learning.}
\label{fig:architecture}
\end{figure}

\subsection{Model Design}
\label{sec:architecture}

Our framework adopts a typical Pairformer--Diffusion architecture~\cite{abramson2024alphafold3}.
As shown in Figure~\ref{fig:architecture}, the input embedder produces initial single $(\mathbf{s}^{\text{init}})$ and pair $(\mathbf{z}^{\text{init}})$ representations, the Pairformer trunk refines them through triangle updates and pair-biased attention, and the diffusion module decodes atomic coordinates conditioned on the refined representations. We augment this pipeline with a \emph{ConstraintEmbedder} that injects fibril-specific geometric priors into $\mathbf{z}^{\text{init}}$.

\paragraph{Constraint-Aware Input Embedding.}
A target fibril complex comprises $K_p$ peptide chains that share an identical amino acid sequence and $K_l$ ligand copies that share an identical molecular graph, which we denote by $\mathbf{A}^{1:K_p}$ and $\mathbf{G}^{1:K_l}$, respectively.
These inputs are tokenised and embedded into initial per-token single representations $\mathbf{s}_i^{\text{init}} \in \mathbb{R}^{c_s}$ and per-token-pair pair representations $\mathbf{z}_{ij}^{\text{init}} \in \mathbb{R}^{c_z}$, where each token $i$ corresponds to either a residue (for protein chains) or a heavy atom (for ligands).
Here $\mathbf{s}_i$ encodes the identity of token $i$, while $\mathbf{z}_{ij}$ encodes the pairwise relationship between tokens $i$ and $j$, including spatial proximity, bonding, and relative position.

To inject fibril-specific geometry into the pair representation, we introduce a \emph{ConstraintEmbedder} that supports two types of constraints.
The \emph{contact constraint} $\mathbf{C}^{\text{ct}}$ captures the translational periodicity of the fibril, i.e., the rigid-body shift of ${\sim}4.8$\,\AA\ between neighbouring chains along the fibril axis.
It is realised as the $\mathrm{C}_\alpha$--$\mathrm{C}_\alpha$ distance $\mathbf{C}^{\text{ct}}_{ij} = \|\mathbf{r}_i - \mathbf{r}_j\|_2$ between residue tokens $i$ and $j$, where $\mathbf{r}_i, \mathbf{r}_j \in \mathbb{R}^{3}$ are reference coordinates taken from the target fibril structure.
The distance is then added to the pair representation through a linear projection:
\begin{equation}
  \mathbf{z}^{\text{init}}_{ij} \;\leftarrow\; \mathbf{z}^{\text{init}}_{ij} + \mathbf{W}_{\text{ct}}\,\mathbf{C}^{\text{ct}}_{ij}.
  \label{eq:constraint_embed}
\end{equation}
The \emph{pocket constraint} $\mathbf{C}^{\text{pkt}}$ marks the target binding region on the fibril surface, giving the ligand a spatial prior over where to bind; it is injected in the same manner via a separate projection $\mathbf{W}_{\text{pkt}}$.
All projections in the ConstraintEmbedder are zero-initialised, so that the added branch does not perturb the pair representation at initialization.

\paragraph{Pairformer Trunk and Diffusion Module.}
\label{sec:fibril_encoding}
The Pairformer trunk serves as the primary representation-learning engine of the model: it progressively refines $(\mathbf{s}^{\text{init}}, \mathbf{z}^{\text{init}})$ through $L$ blocks that enforce triangle-inequality consistency on the pair track and propagate this geometry into the single track via pair-biased attention~\cite{abramson2024alphafold3}.
Because the constraints enter at $\mathbf{z}^{\text{init}}$, they are carried through every Pairformer block and are ultimately encoded into the refined representations $(\mathbf{s}^{(L)}, \mathbf{z}^{(L)})$.
The diffusion module then acts as a generative decoder: conditioned on $(\mathbf{s}^{(L)}, \mathbf{z}^{(L)})$, it parameterises the velocity field $v_\theta(\mathbf{x}_t, t; \mathbf{s}^{(L)}, \mathbf{z}^{(L)})$ introduced in Eq.~\eqref{eq:flow_matching}, and at inference integrates it from $t{=}1$ to $t{=}0$ to denoise random coordinates into a plausible fibril--ligand complex $(\mathbf{x}_\text{prot}, \mathbf{x}_\text{lig})$.

\begin{figure}[t]
\centering
\includegraphics[width=\linewidth]{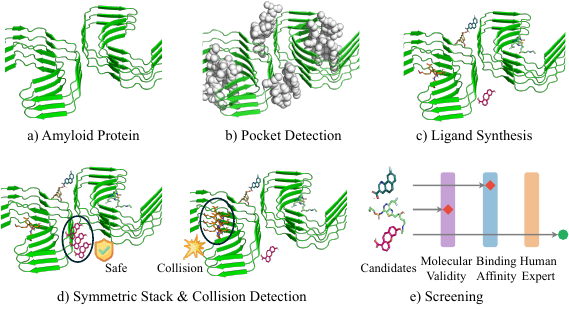}
\caption{Training data curation pipeline. Starting from an amyloid fibril structure (a), we detect candidate binding sites on the fibril surface (b) and generate diverse small-molecule conformers at each site (c). Each conformer is replicated along the fibril axis to form a symmetric stacking arrangement, retaining only collision-free configurations (d). Surviving candidates pass sequential LLM, RDKit, binding-affinity, and expert checks (e).}
\label{fig:data_pipeline}
\end{figure}

\subsection{Training}
\label{sec:training}

\paragraph{Data Curation.}
\label{sec:data_curation}
To address the scarcity of fibril--ligand co-crystal structures, we curate synthetic stacking poses for supervised warm-up (Figure~\ref{fig:data_pipeline}).
We retain two-protofilament fibrils with at least three consecutive chains per protofilament to capture both intra- and inter-protofilament binding interfaces.
For each fibril, fpocket~\cite{le2009fpocket} identifies candidate pockets along the cross-$\beta$ grooves, which are inspected before DrugFlow~\cite{schneuing2025drugflow} generates diverse ligand conformers at each site.
Each conformer is replicated along the fibril axis using the inter-chain translation to mimic the symmetric stacking observed in cryo-EM structures; arrangements with steric clashes or chemically unreasonable conformations are discarded.
Surviving poses pass through sequential filters: coarse LLM-based pose-plausibility screening, RDKit checks of chemical validity, docking-based affinity screening, and final expert structural review. The retained poses form the \textit{AmyDock-Synthetic} warm-up training set.

\paragraph{Warm-up Stage.}
This stage warms up the model to the symmetric binding mode at a geometric level: how ligands intercalate into cross-$\beta$ grooves, align along the fibril axis, and form periodically repeating stacking arrangements.
We supervise the model on the final curated \textit{AmyDock-Synthetic} training set, using a typical structure-prediction training objective~\cite{abramson2024alphafold3} that combines three groups of losses: a flow-matching loss $\mathcal{L}_\text{FM}$ that trains the diffusion module to denoise atomic coordinates towards the ground truth; a distogram loss $\mathcal{L}_\text{dist}$ that supervises the trunk's predicted distribution over pairwise residue distances; and three auxiliary confidence losses ($\mathcal{L}_\text{pLDDT}$, $\mathcal{L}_\text{PDE}$, $\mathcal{L}_\text{PAE}$) that supervise per-atom and per-pair error predictions.
Following the classifier-free guidance strategy~\cite{ho2022classifierfree}, the pocket constraint $\mathbf{C}^{\text{pkt}}$ is dropped with $50\%$ probability during training. This way, the network learns to generate both with and without the pocket prior, and we can combine the two at inference to modulate the strength of the prior.

\paragraph{Reinforcement Learning with Cooperative Binding Reward.}
The warm-up stage captures the overall shape of symmetric stacking, but generated poses still deviate from realistic binding energetics.
To close this gap, we fine-tune the warmed-up model $v_\text{ref}$ with reinforcement learning~\cite{diffusionnft2025} under a composite reward:
\begin{equation}
  r(\mathbf{x}) \;=\; -R_\text{bind}(\mathbf{x}) + \lambda\, R_\text{sym}(\mathbf{x}),
  \label{eq:reward}
\end{equation}
with two complementary terms. The \emph{binding score} combines the AutoDock Vina protein--ligand docking score $\Delta G_\text{PL}$~\cite{trott2010autodock,eberhardt2021autodock} with the ligand--ligand stacking energy $E_\text{stack}$, computed between the centre ligand and its symmetry-generated neighbours using a Mie (8--4) potential with short-distance regularisation:
\begin{equation}
  R_\text{bind} \;=\; \Delta G_\text{PL} + R_\text{stack} \;\equiv\; \Delta G_{\mathrm{eff}}, \qquad R_\text{stack} = f \cdot E_\text{stack},
  \label{eq:dg_eff}
\end{equation}
where $f>0$ weights the stacking contribution. Lower values of this combined score yield higher rewards through the minus sign in Eq.~\eqref{eq:reward}. The \emph{symmetry consistency} $R_\text{sym} = -\,\mathrm{RMSD}(\mathbf{T}\,\mathbf{x}_k,\, \mathbf{x}_{k+1})$ penalises deviations from the inter-chain rigid-body transformation $\mathbf{T}$ (a ${\sim}4.8$\,\AA\ translation along the fibril axis) that should map ligand $\mathbf{x}_k$ at chain $k$ onto $\mathbf{x}_{k+1}$ at chain $k{+}1$. We minimise the RL objective $\mathcal{L}_\text{NFT}(\theta)$ in Eq.~\eqref{eq:nft_loss} with this reward, adding a KL regulariser $\beta_\text{KL}\|v_\theta - v_\text{ref}\|^2$ towards $v_\text{ref}$ to stabilise training.

\section{Experiments}

\subsection{Experimental Setup}
\label{sec:exp_setup}

\paragraph{Implementation.}
We implement the Pairformer--Diffusion backbone using Protenix~\cite{chen2025protenix}, an open-source PyTorch reproduction of AlphaFold~3~\cite{abramson2024alphafold3}.
Training follows the two-stage recipe described in \S\ref{sec:training}.\footnote{See Appendix~\ref{app:data_curation} for data curation, Appendix~\ref{app:reward_expanded} for expanded reward equations, and Appendix~\ref{app:implementation} for implementation details.}
In the warm-up stage, we supervise the model on the final curated \textit{AmyDock-Synthetic} fibril--ligand stacking poses.
In the RL stage, the amyloid fibril--ligand complexes used as roll-out inputs are drawn from the amyloid ligand database of~\citet{chisholm2025ligands} (covering $\alpha$-synuclein, amyloid-$\beta$, and tau fibril targets), filtered to entries with a two-protofilament architecture and excluding evaluation ligands.

\paragraph{Datasets.}
We evaluate primarily on two complementary docking datasets.
\textit{AmyDock-CryoEM} draws ground-truth poses directly from PDB-deposited cryo-EM models; after filtering out solvent, buffer, and other non-specific small molecules, the set comprises 18 fibril--ligand complexes spanning tau, $\alpha$-synuclein, and related fibril-forming proteins.
\textit{AmyDock-Synthetic} generates reference poses through an agent-based sampling pipeline; after plausibility screening, binding-affinity scoring, and expert validation, the held-out evaluation set comprises 57 fibril--ligand pairs that cover a broader range of ligand chemotypes than the cryo-EM structures alone.
Additionally, to probe whether the predicted poses carry signal of the underlying binding energetics, we use a set of 47 small molecules with measured dissociation constants ($K_d$) against $\alpha$-synuclein fibrils published by \citet{chu2015synuclein}. This set has no ground-truth poses and is used solely for binding-affinity rank correlation; we refer to it as \textit{$\alpha$Syn-Bind}.

\paragraph{Baselines.}
We compare against eight docking methods spanning geometry-based docking, regression, diffusion, flow matching, and co-folding paradigms:
PatchDock~\cite{schneidman2005patchdock},
DiffDock~\cite{corso2023diffdock},
DynamicBind~\cite{lu2024dynamicbind},
FABind~\cite{pei2023fabind},
NeuralPLexer~\cite{qiao2024neuralplexer},
FlowDock~\cite{morehead2025flowdock},
Boltz-2~\cite{wohlwend2025boltz2},
and Chai-1~\cite{boitreaud2024chai1}.
For a fair comparison, each learning-based baseline is also fine-tuned on the \textit{AmyDock-Synthetic} training set, as is our approach.

\subsection{Evaluation}

\paragraph{Quantitative Results.}
On \textit{AmyDock-CryoEM} and \textit{AmyDock-Synthetic}, we measure docking accuracy with two metrics: \textbf{Mean / Median RMSD}, the heavy-atom RMSD between the best predicted pose and the ground truth (averaged or median across all pairs); and \textbf{Hit rate ($\leq 5$\,\AA)}, the fraction of pairs with RMSD $\leq 5$\,\AA.
On \textit{$\alpha$Syn-Bind}, we measure binding-affinity correlation with \textbf{Spearman $\rho$} and its $p$-value between predicted binding scores and experimental $\log(K_d/\mathrm{M})$.
For baselines that only support single-ligand docking, we compare the predicted pose against the nearest ground-truth ligand copy.

\begin{table}[!t]
\centering
\caption{Docking accuracy on AmyDock-CryoEM and AmyDock-Synthetic, and binding-affinity rank correlation on $\alpha$Syn-Bind (from~\cite{chu2015synuclein}). RMSD values are in \AA; Hit@5\,\AA\ denotes the fraction of pairs with ligand RMSD $\leq 5$\,\AA; Spearman $\rho$ is computed between predicted binding scores and experimental $\log(K_d/\mathrm{M})$. ``--'' denotes not available (NA).}
\label{tab:main_results}
\small
\setlength{\tabcolsep}{4pt}
\begin{tabular*}{\linewidth}{@{\extracolsep{\fill}}lcccccccc}
\toprule
& \multicolumn{3}{c}{\textbf{AmyDock-CryoEM}} & \multicolumn{3}{c}{\textbf{AmyDock-Synthetic}} & \multicolumn{2}{c}{\textbf{$\alpha$Syn-Bind}} \\
\cmidrule(lr){2-4} \cmidrule(lr){5-7} \cmidrule(lr){8-9}
\textbf{Method} & Mean $\downarrow$ & Median $\downarrow$ & Hit@5\AA\ $\uparrow$ & Mean $\downarrow$ & Median $\downarrow$ & Hit@5\AA\ $\uparrow$ & $\rho$ $\uparrow$ & $p$ $\downarrow$ \\
\midrule
PatchDock         & 23.6  & 24.8  & 0.0\%  & 26.9  & 25.7  & 0.0\%  & --     & -- \\
DiffDock          & 17.67 & 13.68 & 0.0\%  & 19.19 & 19.68 & 3.5\%  & 0.191  & 0.197 \\
DynamicBind       & 26.63 & 29.02 & 0.0\%  & 18.90 & 17.93 & 1.8\%  & 0.102  & 0.544 \\
FABind            & 26.50 & 24.90 & 0.0\%  & 30.34 & 26.87 & 0.0\%  & 0.157  & 0.292 \\
NeuralPLexer      & 17.43 & 19.00 & 5.6\%  & 19.78 & 16.56 & 0.0\%  & $-$0.291 & 0.047 \\
FlowDock          & 19.95 & 21.52 & 0.0\%  & 20.69 & 19.88 & 0.0\%  & 0.148  & 0.322 \\
Boltz-2           & 28.27 & 28.67 & 0.0\%  & 37.13 & 33.87 & 0.0\%  & 0.223  & 0.132 \\
Chai-1            & 19.85 & 19.92 & 0.0\%  & 19.40 & 16.52 & 0.0\%  & $-$0.079 & 0.597 \\
\midrule
\modelname (ours) & \textbf{8.70} & \textbf{4.89} & \textbf{50.0\%} & \textbf{7.35} & \textbf{3.84} & \textbf{68.4\%} & \textbf{0.419} & \textbf{0.004} \\
\bottomrule
\end{tabular*}
\end{table}

As shown in Table~\ref{tab:main_results}, \modelname reaches a hit rate of $\mathbf{68.4\%}$ at $\leq 5$\,\AA\ on \textit{AmyDock-Synthetic} (median RMSD $\mathbf{3.84}$\,\AA) and $\mathbf{50.0\%}$ on \textit{AmyDock-CryoEM}, outperforming baselines whose single-ligand pocket formulations do not explicitly model the cooperative, periodic binding geometry of fibrils. On \textit{$\alpha$Syn-Bind}, \modelname's predicted binding scores show a rank correlation with experimental $K_d$ values (Spearman $\rho{=}\mathbf{0.419}$, $p{=}\mathbf{0.004}$), suggesting potential for ranking ligands by binding affinity.

\begin{figure}[!tb]
\centering
\includegraphics[width=\linewidth]{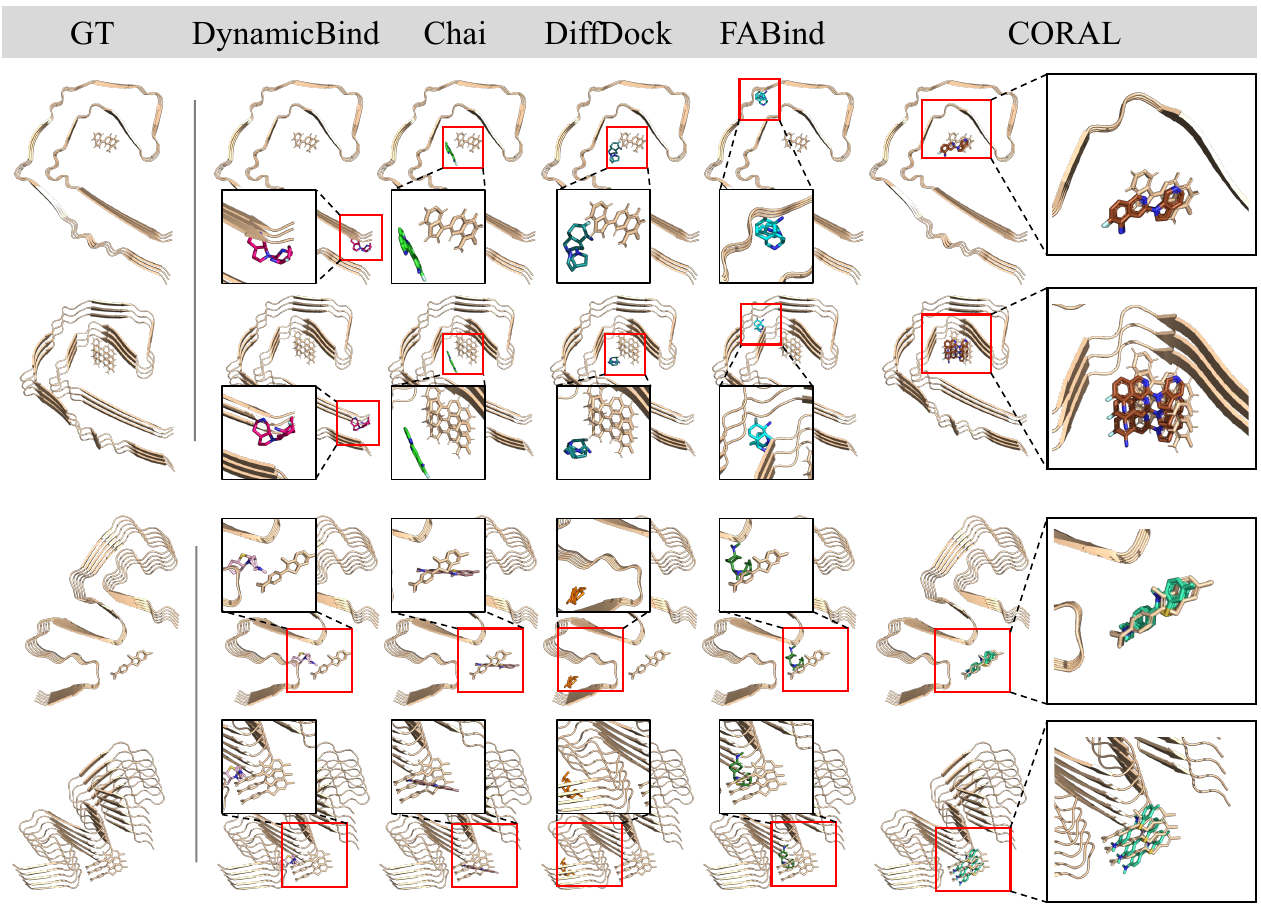}
\caption{Qualitative comparison of predicted fibril--ligand poses on two representative \textit{AmyDock-CryoEM} targets. Each target spans two rows---an axial view along the fibril axis (top) and an oblique top-down view (bottom)---while the columns compare the ground truth against each docking method. The ground-truth ligand is overlaid on every prediction panel; each red inset marks the ligand position that is enlarged in the adjacent black box.}
\label{fig:qualitative}
\end{figure}

\paragraph{Qualitative Results.}
\label{sec:qualitative}
Figure~\ref{fig:qualitative} visualizes predictions on two representative \textit{AmyDock-CryoEM} targets.
Regression-based dockers such as FABind~\cite{pei2023fabind} collapse to a single averaged pose that drifts from the binding region, while the generative baselines dock each ligand independently and, lacking awareness of the fibril's periodic geometry, tend to plaster the ligand vertically against the fibril surface or miss the intra-chain groove entirely---both failure modes yield large RMSD even when the per-ligand chemistry looks plausible.
\modelname instead reproduces the periodic stacking of ligand copies across neighbouring peptide chains.
The visual comparison corroborates the quantitative gap in Table~\ref{tab:main_results} and indicates that learning the inter-ligand cooperative structure (not just single-pose protein--ligand fit) is the key driver of \modelname's improvement.

\FloatBarrier
\subsection{Further Analysis}
\label{sec:further_analysis}

\paragraph{Ablation Study.}
\label{sec:ablation}
We ablate two key design choices of \modelname---the RL reward and the two-stage training recipe---on \textit{AmyDock-Synthetic} (Table~\ref{tab:ablation}).
Each variant modifies a single factor while keeping all other components identical to the full model.
Removing $R_\text{stack}$ collapses $R_\text{bind}$ to plain protein--ligand affinity, eliminating the cooperative inter-ligand signal and degrading $\Delta G_{\mathrm{eff}}$; removing $R_\text{sym}$ has a milder effect on RMSD because the warm-up stage has already established much of the stacking geometry, so $R_\text{sym}$ mainly serves as a fine-tuning anchor during RL; removing the KL term destabilises training and the model diverges from the warmed-up shape prior.
Finally, RL without warm-up converges to a worse local optimum---without a shape prior to start from, the policy explores from scratch and lands on noisier stacking geometry---while warm-up without RL produces plausible stacking but under-performs energetically; only the combined recipe achieves both accurate geometry and favourable binding energetics.

\begin{table}[!tb]
\centering
\caption{Ablation study on AmyDock-Synthetic. Each variant modifies one factor of the full model; ``$-X$'' removes component $X$. $\Delta G_{\mathrm{eff}}$ (kcal/mol) is the combined binding score defined in Eq.~\eqref{eq:dg_eff}; we report both mean and median.}
\label{tab:ablation}
\small
\setlength{\tabcolsep}{4pt}
\begin{tabular*}{\linewidth}{@{\extracolsep{\fill}}lcccc}
\toprule
\textbf{Variant} & \textbf{Mean RMSD $\downarrow$} & \textbf{Median RMSD $\downarrow$} & \textbf{Mean $\Delta G_{\mathrm{eff}}$ $\downarrow$} & \textbf{Median $\Delta G_{\mathrm{eff}}$ $\downarrow$} \\
\midrule
\modelname (full) & \textbf{7.35} & \textbf{3.84} & \textbf{-2.24} & \textbf{-2.44} \\
\midrule
\multicolumn{5}{@{}l}{\textit{RL reward components}} \\
\quad $-\,R_\text{sym}$ (no symmetry term)  & 7.92 & 4.31 & -2.07 & -2.21 \\
\quad $-\,R_\text{stack}$ (no cooperative term)  & 8.21 & 4.46 & -0.78 & -0.95 \\
\quad $-\,$KL (no stability regulariser)    & 12.07 & 8.13 & -1.32 & -1.45 \\
\midrule
\multicolumn{5}{@{}l}{\textit{Training recipe}} \\
\quad Warm-up only (no RL)                   & 8.67 & 5.18 & -1.41 & -1.58 \\
\quad RL only (no warm-up)                   & 13.45 & 9.27 & -0.72 & -0.88 \\
\bottomrule
\end{tabular*}
\end{table}

\begin{figure}[!tbp]
\centering
\includegraphics[width=\linewidth]{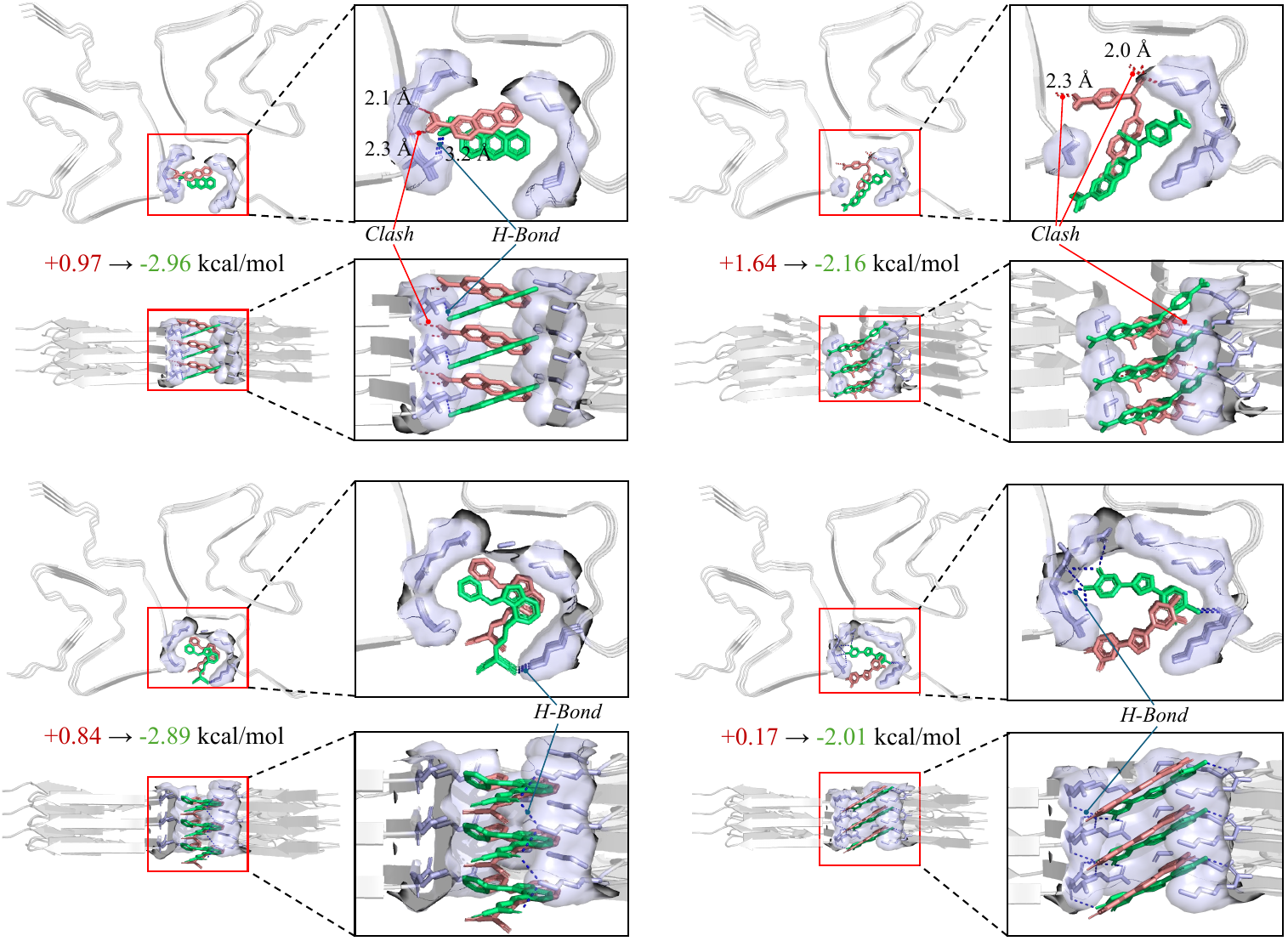}
\caption{\textbf{Pose refinement during reinforcement learning.} Four representative training complexes. For each complex, the pose at its first visit during RL training (\textcolor{brown}{brown}) is shown alongside the pose at its last visit (\textcolor{green!60!black}{green}), with zoomed insets in both top-down (along the fibril axis) and side views; red labels and arrows mark steric clashes (with closest distances in \AA), and blue dashed lines mark hydrogen bonds. Numbers report the corresponding shift in $R_\text{bind}$ from first to last visit, moving from positive (unfavourable) to clearly negative (favourable) values, with $\sim$2--4\,kcal/mol of improvement per complex.}
\label{fig:rl_pose_improvements}
\end{figure}

\paragraph{Pose Refinement during RL.}
During RL, the mean reward stabilises and $R_\text{bind}$ generally becomes more favourable.\footnote{Appendix~\ref{app:training_dynamics} reports training dynamics and the effects of stacking weight on surface burial and interface complementarity.}
Figure~\ref{fig:rl_pose_improvements} illustrates these training-stage changes at the pose level for four representative training complexes, comparing their first and last visits. For each complex, the pose at its first visit during training (brown) fits awkwardly into the cross-$\beta$ groove and exhibits steric clashes with the protein surface. By the last visit (green), the ligand has adapted to the local pocket geometry, and the resulting shape complementarity yields a richer network of favourable protein--ligand interactions: clashes are eliminated, well-defined hydrogen bonds form, and van der Waals packing improves. $R_\text{bind}$ correspondingly shifts from positive (unfavourable) to clearly negative values.

\begin{figure}[!tb]
\centering
\includegraphics[width=\linewidth]{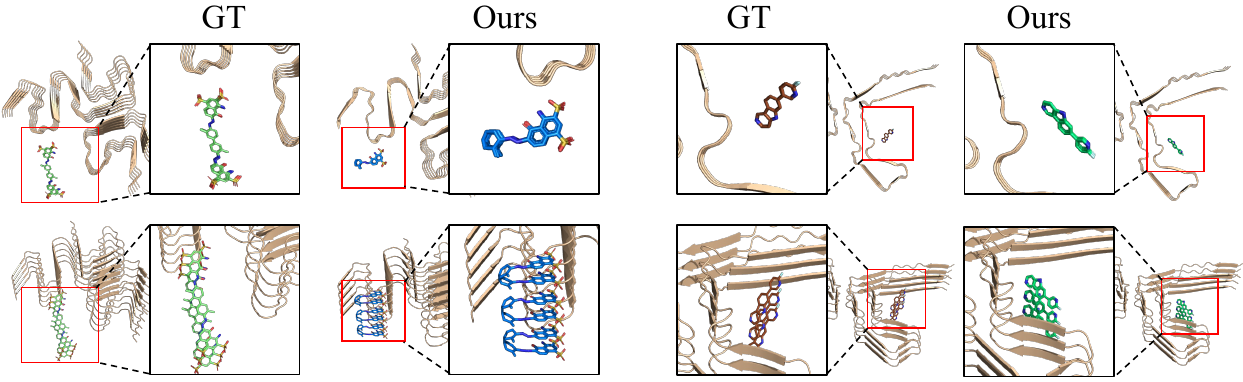}
\caption{Representative failure cases on \textit{AmyDock-CryoEM}. Two cases (columns 1--2 and 3--4) are shown, each pair comparing the ground-truth pose (GT) against \modelname's prediction (Ours) in two views (top row: axial along the fibril axis; bottom row: oblique). Red insets are enlarged in the adjacent panels. The left case shows a prediction with distorted ligand geometry, while the right case shows an orientation mismatch relative to the ground-truth groove site.}
\label{fig:failure}
\end{figure}

\FloatBarrier
\paragraph{Case Study.}
\label{sec:case_study}
We highlight a few cases where \modelname underperforms (Figure~\ref{fig:failure}).
\emph{First}, the model folds the ligand back on itself so that one segment stacks with neighbouring copies while the other bends toward the protein surface (Figure~\ref{fig:failure}, left), letting a single pose gain both stacking and protein-contact energy at once.
\emph{Second}, the model places the ligand at the correct binding site but rotates it sideways from the canonical axial orientation to enlarge protein contact (Figure~\ref{fig:failure}, right).

\FloatBarrier
\section{Conclusion}
\label{sec:conclusion}

We present \modelname, the first deep learning-based docking framework tailored to amyloid fibril--ligand complexes.
Our framework tackles cooperative multi-ligand docking on amyloid fibrils---a binding mode that single-pocket docking models are not designed to capture: it injects fibril-periodicity priors into the structure model and refines pose distributions through online reinforcement learning under a stacking-aware reward, sidestepping the scarcity of fibril--ligand co-crystal supervision.
Across the two docking benchmarks, our framework substantially outperforms existing docking methods. Results on \textit{$\alpha$Syn-Bind} also suggest potential for ranking ligands by binding affinity.

\noindent\textbf{Future Work.}
We plan to explore more accurate numerical simulations, such as molecular dynamics, to develop reward signals that better reflect realistic fibril--ligand binding modes. We also aim to extend our evaluation to other amyloid fibrils, including those formed by transthyretin and immunoglobulin light chains, and to integrate \emph{de novo} ligand generation, our docking framework, and downstream filtering into a single agent-orchestrated closed-loop pipeline~\cite{lu2026aiscientist,boiko2023coscientist} for automated discovery of fibril-binding candidates with minimal human intervention.

\noindent\textbf{Limitations.}
Our evaluation covers periodic ligand-stacking geometries in tau, $\alpha$-synuclein, and amyloid-$\beta$ fibrils. The synthetic data pipeline may introduce biases during generation and screening, with its LLM pre-screen relying on prompt-based judgments. Equipping this screening agent with established structural and chemical analysis tools could improve reliability. The training reward approximates intermolecular interactions through a docking score and a pairwise Mie (8--4) stacking potential. The geometric distortions and orientation mismatches in Figure~\ref{fig:failure} motivate a more complete treatment of conformational and intermolecular energetics. Our affinity results suggest potential for ligand ranking; broader experimental binding datasets and more accurate numerical models of fibril--ligand interactions could support a more comprehensive assessment and refinement of affinity prediction.

\noindent\textbf{Broader Impact.}
The framework aims to accelerate early-stage drug discovery for Alzheimer's, Parkinson's, and related neurodegenerative diseases. Predictions of cooperative fibril--ligand binding provide structural hypotheses that can help prioritise candidate imaging probes and therapeutic ligands for experimental evaluation.

\begin{ack}
The research reported in this publication was supported by a gift from Google, administered through the King Abdullah University of Science and Technology (KAUST) -- Center of Excellence for Generative AI under award number 5940.
We also thank KAUST for providing GPU computing resources on the Ibex cluster.
The authors declare no competing interests.
\end{ack}

\bibliographystyle{abbrvnat}
\bibliography{refs}

\newpage
\appendix

{\centering\Large\bfseries Appendix\par}
\vspace{0.8em}

\noindent
This appendix provides additional details supporting the main paper.
\S\ref{app:architecture} describes the architecture, supervised warm-up, the synthetic data-curation pipeline used to construct \textit{AmyDock-Synthetic}, and the \textit{AmyDock} benchmark.
\S\ref{app:rl_details} presents the full reinforcement-learning algorithm, the cooperative binding reward design, and the pairwise stacking potential.
\S\ref{app:more_qual} provides additional qualitative visualisations of \modelname's predicted fibril--ligand poses.

\section{Warm-up and Data Curation}
\label{app:architecture}

This section provides additional details on the warm-up training stage and the synthetic data-curation pipeline used to construct \textit{AmyDock-Synthetic}. Because amyloid fibril--ligand co-crystal structures are scarce, we curate a synthetic dataset of symmetric stacking poses via an agent-driven pipeline (\S\ref{app:data_curation}), and warm up the model on this dataset to first establish the symmetric binding mode at a geometric level (\S\ref{app:warmup_training}) before reinforcement-learning fine-tuning toward cooperative binding energetics (Appendix~\ref{app:rl_details}).

\subsection{Initialisation and Warm-up Training}
\label{app:warmup_training}

\paragraph{Initialisation and Generic Docking Pre-Training.}
We train the Protenix Pairformer--Diffusion backbone~\cite{chen2025protenix} from random initialisation on a diverse collection of protein--ligand co-crystals using the supervised objective defined below. This generic docking pre-training combines PLINDER~\cite{durairaj2024plinder}, Binding MOAD~\cite{hu2005bindingmoad}, HiQBind~\cite{wang2025hiqbind}, and SPINDR~\cite{cremer2025flowr} to expose the model to a broad range of small-molecule geometries and chemotypes. Amyloid-specific warm-up then teaches periodic stacking geometry, followed by RL refinement of cooperative binding energetics. The added ConstraintEmbedder projections are zero-initialised as described below.

\paragraph{Warm-up Objective.}
For the warm-up stage, we adopt the standard AlphaFold-3-style structure-prediction objective~\cite{abramson2024alphafold3,chen2025protenix}, which combines three groups of losses: (i) a \emph{diffusion} loss $\mathcal{L}_\text{diff}$ that supervises atomic coordinates (instantiating $\mathcal{L}_\text{FM}$ in Eq.~\eqref{eq:flow_matching}), (ii) a \emph{distogram} loss $\mathcal{L}_\text{dist}$ that supervises pairwise residue distances, and (iii) a \emph{confidence} loss $\mathcal{L}_\text{conf}$ that trains the model to predict its own per-atom and per-pair errors. The total warm-up objective is
\begin{equation}
  \mathcal{L}_\text{SFT} \;=\; \alpha_\text{diff}\,\mathcal{L}_\text{diff} \;+\; \alpha_\text{dist}\,\mathcal{L}_\text{dist} \;+\; \alpha_\text{conf}\,\mathcal{L}_\text{conf},
  \label{eq:sft_total_loss}
\end{equation}
with each component defined as
\begin{align}
  \mathcal{L}_\text{diff} &= \mathcal{L}_\text{MSE} + \alpha_\text{bond}\,\mathcal{L}_\text{bond} + w_\text{lddt}\,\mathcal{L}_\text{smooth-LDDT}, \label{eq:sft_loss_diff} \\
  \mathcal{L}_\text{conf} &= \mathcal{L}_\text{pLDDT} + \mathcal{L}_\text{PDE} + \alpha_\text{pae}\,\mathcal{L}_\text{PAE}. \label{eq:sft_loss_conf}
\end{align}
Here $\mathcal{L}_\text{MSE}$ is the weighted per-atom coordinate MSE (with element-specific weights $w_\text{protein}{=}1$, $w_\text{DNA}{=}w_\text{RNA}{=}5$, $w_\text{ligand}{=}10$), $\mathcal{L}_\text{bond}$ is a bond-length consistency loss, $\mathcal{L}_\text{smooth-LDDT}$ is the smooth-LDDT loss, $\mathcal{L}_\text{dist}$ is the distogram cross-entropy over binned $\text{C}_\beta$--$\text{C}_\beta$ distances, and $\mathcal{L}_\text{pLDDT}, \mathcal{L}_\text{PDE}, \mathcal{L}_\text{PAE}$ are the three confidence-head classification losses. The global weights are empirically set as $\alpha_\text{diff}{=}4.0$, $\alpha_\text{dist}{=}0.03$, $\alpha_\text{conf}{=}10^{-4}$, $w_\text{lddt}{=}1.0$, while $\alpha_\text{bond}$ and $\alpha_\text{pae}$ act as switches that are set to $0$ during early training and turned on ($=1$) in the final fine-tuning stage.

\paragraph{Geometry Constraint.}
\label{app:constraints}
Beyond the loss design, we inject two fibril-specific priors into the pair representation $\mathbf{z}^{\text{init}}$ via the ConstraintEmbedder (\S\ref{sec:architecture}). The \emph{contact constraint} $\mathbf{C}^{\text{ct}}$ encodes the translational periodicity of cross-$\beta$ amyloid fibrils, which are stacks of identical $\beta$-strand layers separated by a characteristic inter-strand C$_\alpha$--C$_\alpha$ spacing of ${\sim}4.8$\,\AA. For each pair of residue tokens $(i, j)$ we compute the pairwise distance $\mathbf{C}^{\text{ct}}_{ij} = \|\mathbf{r}_i - \mathbf{r}_j\|_2$ from the reference fibril structure and inject it through a learned linear projection $\mathbf{W}_{\text{ct}}$ (Eq.~\ref{eq:constraint_embed}); the constraint then propagates to every Pairformer block, biasing attention toward the periodic ladder of inter-chain contacts characteristic of cross-$\beta$ assemblies.
The \emph{pocket constraint} $\mathbf{C}^{\text{pkt}}$ marks the candidate binding region on the fibril surface (the residues bordering a cross-$\beta$ groove), providing a soft spatial prior over where ligands should bind. It is injected with the same mechanism through a separate projection $\mathbf{W}_{\text{pkt}}$. Both projections are zero-initialised so that the constraint branch does not perturb the pair representation at the start of training. Following classifier-free guidance~\cite{ho2022classifierfree}, $\mathbf{C}^{\text{pkt}}$ is dropped with probability $0.5$ during warm-up; the model thus learns both with and without the pocket prior, and the prior strength can be modulated at inference by interpolating between the conditional and unconditional velocity fields.

\subsection{Data Curation}
\label{app:data_pipeline}
\label{app:data_curation}

\begin{figure}[H]
\centering
\includegraphics[width=\linewidth]{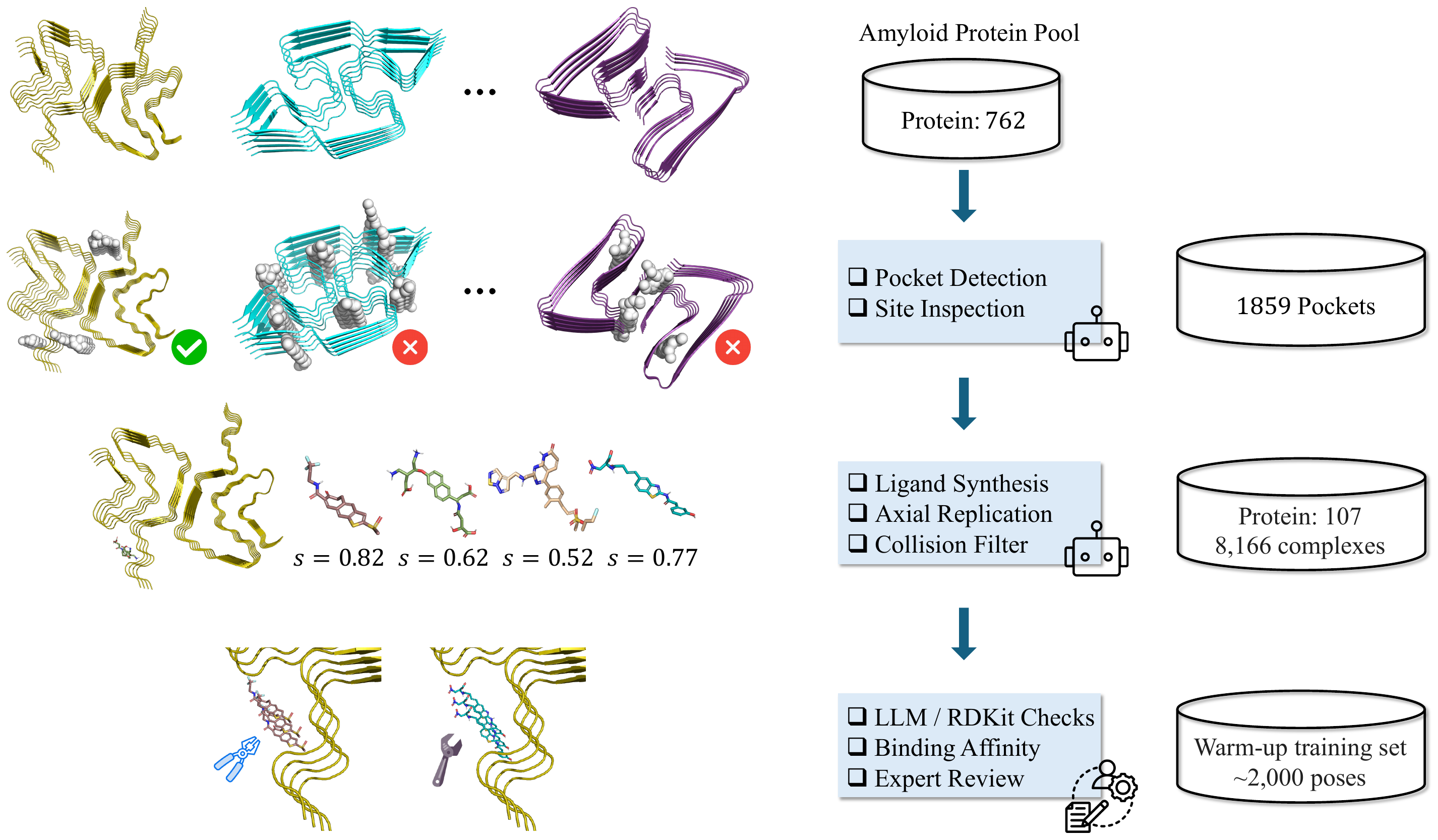}
\caption{Data curation for the \textit{AmyDock-Synthetic} warm-up training set. Pocket detection and inspection precede ligand generation, axial replication, and collision filtering. LLM/RDKit checks, docking-based affinity screening, and expert review yield approximately 2{,}000 training poses; $s\in[0,1]$ denotes the LLM pose-plausibility screening score.}
\label{fig:sup_data_pipeline}
\end{figure}

Resolved co-crystal structures of amyloid fibril--ligand complexes are exceptionally scarce: even with recent cryo-EM advances, only a handful have been structurally characterised, making supervised training of fibril-specific docking models infeasible from experimental data alone. To overcome this bottleneck, we synthesise plausible cooperative stacking poses through an agent-driven pipeline (Figure~\ref{fig:sup_data_pipeline}) and validate them via geometric, LLM-based, and expert checks, yielding the \textit{AmyDock-Synthetic} training set used in the warm-up stage. The remainder of this subsection describes each stage in detail.

\paragraph{Curation Workflow.}
\label{app:data_stats}
The workflow comprises three sequential stages: (1) pocket detection and site inspection; (2) ligand generation, axial replication, and collision filtering; and (3) LLM-based plausibility screening, RDKit validation, docking-based affinity screening, and expert structural review. We detail these operations and their intermediate outputs below.

\textit{Pocket detection and site inspection.} We identify candidate binding sites on the fibril surface using two complementary tools, Fpocket~\cite{le2009fpocket} and CaverDock~\cite{filipovic2019caverdock}, and merge their outputs followed by manual expert inspection to discard non-physical or duplicate sites. This yields \textbf{1{,}859 candidate pockets} across all fibril targets.

\textit{Conformer generation.} For each pocket, we generate \textbf{100 candidate ligands} using a fine-tuned variant of DrugFlow~\cite{schneuing2025drugflow}; we fine-tune DrugFlow on training targets that contain $\beta$-sheet structures, biasing the generative distribution toward chemotypes more compatible with cross-$\beta$ groove binding. This stage takes approximately \textbf{one week on 4~NVIDIA A100 GPUs}.

\textit{Symmetric stacking and collision filtering.} Each conformer is replicated along the fibril axis and tested for inter-ligand and ligand--protein steric clashes (atom-pair distance $<2.0$\,\AA). After this filter, \textbf{8{,}166 candidate stacking arrangements} remain.

\textit{Pose screening and expert review.} Surviving arrangements undergo coarse LLM-based screening for binding-mode plausibility (prompts in \S\ref{app:llm_prompts}) and chemical-validity checks with RDKit's sanitisation routines. Candidates that pass these checks undergo docking-based affinity screening followed by expert structural review; experts additionally adjust ambiguous cases. The resulting \textit{AmyDock-Synthetic} warm-up training set contains approximately \textbf{2{,}000 poses}.

\paragraph{LLM-Based Pose Screening.}
\label{app:llm_prompts}

GPT-5 provides a qualitative plausibility screen for the stacking poses produced by DrugFlow and axial replication. Given a PDB excerpt of a fibril--ligand complex, it assesses whether the ligand occupies a groove, aligns with the fibril axis, and makes plausible contacts. This screen removes obvious binding-mode mismatches before the remaining chemical, affinity, and expert assessments. The model returns a binary verdict and a screening score in $[0,1]$ using the following prompts.

\begin{promptbox}[System Prompt]
You are an expert structural biologist and computational chemist specializing in amyloid fibril structures and small molecule binding. Your task is to evaluate whether a small molecule is properly docked in the grooves/channels of a fibril protein structure. Fibril proteins have characteristic cross-$\beta$ sheet structures with grooves or channels running along the fibril axis where small molecules can bind.

\vspace{4pt}
Key characteristics of proper fibril groove docking:\\
(1) Location: the molecule should be positioned in the groove/channel between protein chains, not on the surface or outside the fibril.\\
(2) Orientation: the molecule should align along the fibril axis, not perpendicular to it.\\
(3) Interactions: favorable contacts with groove walls, including Van der Waals contacts (3.0--5.0 \AA), hydrogen bonds (2.5--3.5 \AA), $\pi$-$\pi$ stacking (3.3--4.5 \AA) where applicable, and hydrophobic contacts.\\
(4) Shape complementarity: the molecule should fit well into the groove geometry.\\
(5) No clashes: no severe steric clashes ($<$2.0 \AA).
\end{promptbox}

\begin{promptbox}[User Prompt]
I have a PDB structure file containing a fibril protein with a docked small molecule. Please evaluate whether the small molecule is properly docked in the fibril groove/channel.

\vspace{4pt}
PDB File Content: \textcolor{blue!70!black}{\{pdb\_content\}}

\vspace{4pt}
Evaluation Criteria:\\
(1) Location: is the molecule positioned in a groove/channel between protein chains (not on surface)?\\
(2) Orientation: is the molecule aligned along the fibril axis?\\
(3) Interactions: does the molecule form favorable interactions with groove residues (VdW, H-bond, $\pi$-$\pi$, hydrophobic)?\\
(4) Shape complementarity: does the molecule fit well into the groove geometry?\\
(5) No clashes: no severe steric clashes ($<$2.0 \AA).

\vspace{4pt}
Output Format:\\
Answer: [Yes or No]\\
Score: [number between 0.0 and 1.0]
\end{promptbox}

Candidates are retained when the LLM returns a positive verdict and a screening score of at least $0.5$; the score is not used for ranking. It expresses qualitative pose plausibility rather than a physical energy, binding affinity, or calibrated probability. Axial replication, collision filtering, RDKit sanitisation, affinity screening, and expert review provide the complementary geometric and chemical assessment.

\paragraph{AmyDock Training and Evaluation Sets.}
\label{app:amydock_details}
The approximately 2{,}000 curated \textit{AmyDock-Synthetic} poses produced by this pipeline are used for warm-up training. The \textit{AmyDock-Synthetic} evaluation benchmark comprises a separate held-out set of 57 fibril--ligand pairs. Each evaluation pose undergoes stricter validation than the warm-up training poses: a more stringent binding-affinity criterion, an energy-minimisation check requiring a small heavy-atom RMSD change between the initial and relaxed poses, and manual expert structural review.
Both the synthetic evaluation set and the experimentally-resolved \textit{AmyDock-CryoEM} set consist of two-protofilament fibrils with at least three consecutive chains per protofilament, covering both intra- and inter-protofilament binding interfaces.

\begin{samepage}
\paragraph{Ligand Scaffold Overlap.}
To assess chemical overlap between the training and evaluation sets beyond exact ligand identity, we compare Bemis--Murcko scaffolds~\cite{bemis1996properties}, which represent a molecule's ring systems and connecting linkers. Amyloid binders often share extended, planar aromatic motifs, so fingerprint similarity can remain high across this ligand class. Scaffold comparison therefore examines whether ligands share the same core molecular framework. We count an overlap when an evaluation ligand has the same scaffold as any training ligand. Every evaluation ligand in \textit{AmyDock-CryoEM} and \textit{AmyDock-Synthetic} has a Bemis--Murcko scaffold absent from the training set, yielding zero scaffold overlap for both benchmarks. This separation concerns core scaffolds; local chemical motifs may still be shared.

\end{samepage}

\subsection{Implementation Details}
\label{app:implementation}

Table~\ref{tab:architecture_hparams} summarises the backbone dimensions, sampling settings, and optimisation parameters.
The warm-up stage uses a learning rate of $1\times 10^{-4}$, while the RL stage uses a lower learning rate of $3\times 10^{-6}$ for training stability.
Both stages are trained on $8$ NVIDIA A100 80G GPUs; the warm-up stage takes approximately $3$ days and the RL stage approximately $7$ days.
The weight EMA with decay $0.999$ is distinct from the sampling-policy EMA in \S\ref{app:nft_training_objective}. Additional RL hyperparameters are given in Table~\ref{tab:nft_hparams}.

\begingroup
\small
\renewcommand{\arraystretch}{1.05}
\setlength{\LTcapwidth}{\linewidth}
\begin{xltabular}{\linewidth}{@{}lX@{}}
\caption{Model architecture and optimisation settings. Stage-specific values are labelled explicitly.}
\label{tab:architecture_hparams}\\
\toprule
\textbf{Hyperparameter} & \textbf{Value} \\
\midrule
\endfirsthead
\caption[]{Model architecture and optimisation settings (continued).}\\
\toprule
\textbf{Hyperparameter} & \textbf{Value} \\
\midrule
\endhead
\midrule
\multicolumn{2}{r@{}}{\textit{Continued on next page}} \\
\endfoot
\bottomrule
\endlastfoot
\multicolumn{2}{@{}l}{\textit{Representations}} \\*
Single (token) width & 384 \\
Pair width & 128 \\
Atom / atom-pair width & 128 / 16 \\
\midrule
\multicolumn{2}{@{}l}{\textit{Pairformer trunk}} \\*
Blocks & 48 \\
Attention heads & 16 \\
Dropout & 0.25 \\
Recycling iterations & 10 \\
\midrule
\multicolumn{2}{@{}l}{\textit{Diffusion module}} \\*
Token width & 768 \\
Atom encoder / decoder blocks & 3 / 3 \\
Atom encoder / decoder heads & 4 / 4 \\
Token transformer blocks & 24 \\
Token transformer heads & 16 \\
Sampling steps (training roll-outs) & 20 \\
Sampling steps (inference) & 200 \\
\midrule
\multicolumn{2}{@{}l}{\textit{Heads and embedders}} \\*
Distogram bins & 64 \\
ConstraintEmbedder & Contact and pocket projections, zero-initialised \\
MSA module / template embedder & Disabled / disabled \\
\midrule
\multicolumn{2}{@{}l}{\textit{Optimisation}} \\*
Optimiser & Adam ($\beta_1=0.9$, $\beta_2=0.95$) \\*
Weight decay & $10^{-8}$ \\*
Learning rate (warm-up / RL) & $1\times10^{-4}$ / $3\times10^{-6}$ \\*
Gradient clipping (warm-up / RL) & 10 / 1 \\*
Weight EMA decay & 0.999 \\*
Crop size (warm-up) & 384 tokens \\*
Roll-outs per complex $G$ & 24 \\
\end{xltabular}
\endgroup

\section{Reinforcement Learning with Cooperative Reward}
\label{app:rl_details}
\label{app:nft}

This section details the reinforcement-learning stage of \modelname\ used in \S\ref{sec:training}. We first present the training objective based on DiffusionNFT~\cite{diffusionnft2025} (\S\ref{app:nft_training_objective}), then describe how the composite reward combines protein--ligand docking, ligand--ligand stacking modelled by a Mie (8--4) potential, and symmetry consistency (\S\ref{app:reward_expanded}).

\subsection{Training Objective}
\label{app:nft_training_objective}

DiffusionNFT~\cite{diffusionnft2025} is an online reinforcement-learning algorithm for diffusion (and flow-matching) models that operates directly on the forward process and avoids explicit likelihood estimation.

\paragraph{Forward Process and Sampling.}
Given a clean sample $\mathbf{x}_0 \sim p_{\text{data}}$ and a noise sample $\boldsymbol{\epsilon} \sim \mathcal{N}(\mathbf{0}, \mathbf{I})$, the \emph{forward process} interpolates between data and noise according to a schedule $(\alpha_t, \sigma_t)$:
\begin{equation}
  \mathbf{x}_t \;=\; \alpha_t \mathbf{x}_0 + \sigma_t \boldsymbol{\epsilon}, \qquad t \in [0, 1],
\end{equation}
with target velocity
\begin{equation}
  \mathbf{v}(\mathbf{x}_t, t) \;=\; \dot{\alpha}_t \mathbf{x}_0 + \dot{\sigma}_t \boldsymbol{\epsilon}.
  \label{eq:fm_target}
\end{equation}
Standard flow matching trains a velocity field $v_\theta$ by $\mathcal{L}_\text{FM}(\theta) = \mathbb{E}\,\|v_\theta(\mathbf{x}_t, t) - \mathbf{v}\|_2^2$. At inference, samples are generated by integrating the learned velocity field along the reverse direction: starting from $\mathbf{x}_1 \sim \mathcal{N}(\mathbf{0}, \mathbf{I})$ and solving $\mathrm{d}\mathbf{x}_t / \mathrm{d}t = v_\theta(\mathbf{x}_t, t)$ from $t=1$ to $t=0$ to obtain a clean sample $\mathbf{x}_0$.

\paragraph{Implicit Positive / Negative Policies.}
Given a frozen sampling policy $v_\text{old}$ and the current policy $v_\theta$, DiffusionNFT defines two \emph{implicit} velocity fields through reparameterisation:
\begin{align}
  v_\theta^{+}(\mathbf{x}_t, t) &:= (1 - \beta)\,v_\text{old}(\mathbf{x}_t, t) + \beta\,v_\theta(\mathbf{x}_t, t), \label{eq:nft_vplus}\\
  v_\theta^{-}(\mathbf{x}_t, t) &:= (1 + \beta)\,v_\text{old}(\mathbf{x}_t, t) - \beta\,v_\theta(\mathbf{x}_t, t), \label{eq:nft_vminus}
\end{align}
where $\beta \in (0, 1)$ is a mixing coefficient. Here $v_\theta^{+}$ is the \emph{positive} policy (interpolating from $v_\text{old}$ towards $v_\theta$) and $v_\theta^{-}$ is the \emph{negative} policy (extrapolating away from $v_\theta$); both are fully determined by $v_\theta$, with no auxiliary guidance networks.

\paragraph{Reward-Weighted Loss.}
Let $\tilde r(\mathbf{x}_0) \in [0, 1]$ denote the \emph{normalised reward weight} of a sampled clean trajectory endpoint $\mathbf{x}_0$ (we describe how we obtain $\tilde r$ from the raw composite reward $r$ in the next paragraph). The NFT loss minimised at each optimisation step is
\begin{equation}
  \mathcal{L}_\text{NFT}(\theta) \;=\; \mathbb{E}\!\left[\, \tilde r(\mathbf{x}_0)\,\|v_\theta^{+}(\mathbf{x}_t, t) - \mathbf{v}\|_2^2 \;+\; (1 - \tilde r(\mathbf{x}_0))\,\|v_\theta^{-}(\mathbf{x}_t, t) - \mathbf{v}\|_2^2 \,\right],
  \label{eq:nft_loss_full}
\end{equation}
with the expectation taken over $(\mathbf{x}_0, \boldsymbol{\epsilon}, t)$. High-reward samples are regressed by the \emph{positive} branch $v_\theta^{+}$ and low-reward samples by the \emph{negative} branch $v_\theta^{-}$, so the policy is simultaneously pushed towards $v_\theta^{+}$ and away from $v_\theta^{-}$. We refer to~\cite{diffusionnft2025} for the gradient analysis showing this corresponds to a soft importance-weighted policy improvement without explicit likelihood ratios.

\paragraph{Reward Normalisation.}
\label{app:nft_reward_map}
Our composite reward $r(\mathbf{x}) = -R_\text{bind}(\mathbf{x}) + \lambda\,R_\text{sym}(\mathbf{x})$ (Eq.~\eqref{eq:reward}) assigns higher values to more favourable poses. Group-relative normalisation maps this reward to the weight $\tilde r \in [0, 1]$ used in the NFT loss, comparing $G$ on-policy rollouts $\{\mathbf{x}^{(i)}\}_{i=1}^{G}$ sampled from the same complex:
\begin{equation}
  \tilde r^{(i)} \;=\; \frac{r(\mathbf{x}^{(i)}) - \min_j r(\mathbf{x}^{(j)})}{\max_j r(\mathbf{x}^{(j)}) - \min_j r(\mathbf{x}^{(j)}) + \varepsilon}.
  \label{eq:reward_norm}
\end{equation}
This increasing map preserves the reward ordering, assigning larger weights to higher-reward rollouts and hence greater emphasis to the positive branch $v_\theta^{+}$. The small constant $\varepsilon = 10^{-6}$ prevents division by zero when all rewards are equal.

\paragraph{Sampling Policy Update.}
To keep the on-policy assumption approximately valid while preventing $v_\text{old}$ from lagging too far behind $v_\theta$, we update $v_\text{old}$ by a soft exponential moving average (EMA) after every $N_\text{upd}$ gradient steps:
\begin{equation}
  \theta_\text{old} \;\leftarrow\; \eta\,\theta_\text{old} + (1 - \eta)\,\theta,
  \label{eq:ema}
\end{equation}
with $\eta \in [0, 1]$ the EMA coefficient.

\paragraph{KL Regularisation.}
To preserve the stacking geometry learned during warm-up, we regularise the RL policy towards the frozen warm-up model $v_\text{ref}$:
\begin{equation}
  \mathcal{L}_\text{KL}(\theta) \;=\; \mathbb{E}_{\mathbf{x}_t, t}\!\left[\, \|v_\theta(\mathbf{x}_t, t) - v_\text{ref}(\mathbf{x}_t, t)\|_2^2 \,\right],
\end{equation}
and optimise the combined objective $\mathcal{L}_\text{NFT}(\theta) + \beta_\text{KL}\,\mathcal{L}_\text{KL}(\theta)$. This velocity-space penalty acts as a proximal regulariser rather than an explicit KL divergence.

\paragraph{Hyperparameters.}
The mixing coefficient $\beta$, EMA coefficient $\eta$, and group size $G$ follow the defaults of DiffusionNFT~\cite{diffusionnft2025}. We update $v_\text{old}$ every $N_\text{upd}{=}8$ epochs (rather than every step) to give the behaviour policy a more stable optimisation window on our relatively small cooperative-binding training set, and increase the KL weight from the DiffusionNFT default of $10^{-3}$ to $\beta_\text{KL}{=}10^{-2}$ to constrain policy drift from the warmed-up reference. The symmetry-reward weight $\lambda$ was tuned on a held-out subset of fibrils.
\begin{table}[h]
\centering
\caption{DiffusionNFT hyperparameters used in the RL stage.}
\label{tab:nft_hparams}
\vspace{0.5em}
\begin{tabularx}{\linewidth}{@{}lcX@{}}
\toprule
\textbf{Symbol} & \textbf{Value} & \textbf{Description} \\
\midrule
$\beta$                & $0.1$    & Mixing coefficient in $v_\theta^{\pm}$ (Eq.~\ref{eq:nft_vplus}--\ref{eq:nft_vminus}) \\
$\lambda$              & $0.01$   & Weight of $R_\text{sym}$ in the composite reward \\
$\beta_\text{KL}$      & $0.01$   & KL regularisation weight \\
$\eta$                 & $0.5$    & EMA coefficient for $v_\text{old}$ (Eq.~\ref{eq:ema}) \\
$N_\text{upd}$         & $8$      & Gradient steps between $v_\text{old}$ updates \\
$G$                    & $24$     & Rollouts per complex for reward normalisation (Eq.~\ref{eq:reward_norm}) \\
\bottomrule
\end{tabularx}
\end{table}

\subsection{Reward Design}
\label{app:reward_expanded}

The composite reward $r(\mathbf{x}) = -R_\text{bind}(\mathbf{x}) + \lambda\, R_\text{sym}(\mathbf{x})$ (\S\ref{sec:training}) favours energetically compatible ligand stacks with consistent geometry across neighbouring binding sites. The binding term $R_\text{bind}$ combines protein--ligand docking and ligand--ligand packing, while $R_\text{sym}$ penalises deviations from the fibril's repeating geometry. With $\lambda>0$, maximising $r$ favours lower binding scores and smaller symmetry deviations.

\paragraph{Binding Score.}
Small-molecule ligands that bind amyloid fibrils occupy periodic binding sites along the cross-$\beta$ groove, where adjacent ligands are separated by the inter-strand distance (${\sim}4.7$--$4.9$\,\AA). This periodic arrangement gives rise to cooperative ligand--ligand interactions that are not captured by standard protein--ligand docking scores~\cite{smith2024symdock,smith2025cooperative}. We therefore augment the AutoDock Vina protein--ligand affinity with a stacking-energy correction:
\begin{equation}
  R_\text{bind} \;=\; \Delta G_\text{PL} + f \cdot E_\text{stack},
  \label{eq:rbind_app}
\end{equation}
where $\Delta G_\text{PL}$ is the empirical Vina docking score (kcal/mol)~\cite{trott2010autodock,eberhardt2021autodock}, and $E_\text{stack}$ sums Mie (8--4) pair interactions with short-distance regularisation between the centre ligand and its neighbouring copies. We use $\Delta G_{\mathrm{eff}}$ as shorthand for this combined binding score. It provides an approximate energetic objective for pose optimisation and ligand ranking, rather than a calibrated estimate of solution-phase binding free energy.

\paragraph{Ligand--Ligand Stacking Potential.}
We model inter-ligand packing using a Mie (8--4) potential with short-distance regularisation. For an atom pair $(i,j)$ separated by distance $r$, let $d_{ij}=R_i+R_j$ denote its preferred contact distance, with $R_i$ and $R_j$ the assigned atomic contact radii. The regularised pair potential is
\begin{equation}
  E_{\mathrm{Mie}}(r;d_{ij})
  = \varepsilon\left[
    \left(\frac{d_{ij}}{\bar r}\right)^8
    - 2\left(\frac{d_{ij}}{\bar r}\right)^4
  \right],
  \qquad \bar r=\max(r,s\,d_{ij}),
  \label{eq:mie84}
\end{equation}
where $\varepsilon>0$ sets the well depth and $s\in(0,1)$ specifies a short-distance floor that keeps the energy finite for overlapping atoms. The underlying Mie (8--4) potential is a generalised Lennard--Jones potential: the eighth-power term provides short-range repulsion and the fourth-power term provides attraction, with a minimum at $r=d_{ij}$ and energy $-\varepsilon$. The distance floor regularises the short-range interaction while preserving this preferred contact distance.

Let $A$ contain the heavy atoms of the centre ligand and $B_k$ those of neighbouring copy $k$, generated by the fibril's inter-chain symmetry operation. The stacking contribution is
\begin{equation}
  E_\text{stack}
  = \sum_{k\in\mathcal N}\sum_{i\in A}\sum_{j\in B_k}
    \mathbf{1}\!\left[r_{ij}^{(k)}<r_c\right]\,
    E_{\mathrm{Mie}}\!\left(r_{ij}^{(k)};d_{ij}\right),
  \label{eq:stack_sum}
\end{equation}
where $\mathcal N$ is the set of neighbouring ligand copies, $r_{ij}^{(k)}$ is the corresponding atom-pair distance, and $r_c$ is the pairwise cutoff. This stacking term contains no separate Coulomb contribution.

\paragraph{Stacking Weight.}
The empirical weight $f=0.6$ balances ligand--ligand packing against the protein--ligand docking score. Its effective strength depends on the well depth $\varepsilon$ and the atom-pair summation: for fixed geometry and contact radii, the weighted stacking term scales with $f\varepsilon$. This term models local packing interactions; explicit solvation and configurational entropy are outside its scope.

\paragraph{Symmetry Consistency.}
The second reward term encourages ligand poses at neighbouring sites to follow the protein's repeating geometry. Let $\mathbf{T} = (\mathbf{R}, \mathbf{t})$ map chain $k$ onto its nearest neighbour $k{+}1$ within the same protofilament, corresponding to a translation of approximately $4.8$\,\AA\ along the fibril axis. We estimate $\mathbf{T}$ by Kabsch alignment~\cite{kabsch1976solution} of the corresponding C$_\alpha$ atoms. For ligand poses $\mathbf{x}_k$ and $\mathbf{x}_{k+1}$ at these adjacent sites, symmetry consistency is measured by:
\begin{equation}
  R_\text{sym} \;=\; -\,\mathrm{RMSD}\!\left(\mathbf{T}\,\mathbf{x}_k,\; \mathbf{x}_{k+1}\right).
\end{equation}

\subsection{Training Dynamics and Stacking-Weight Sensitivity}
\label{app:training_dynamics}

\begin{figure}[!htbp]
\centering
\begin{minipage}{0.48\linewidth}
\centering
\includegraphics[width=\linewidth]{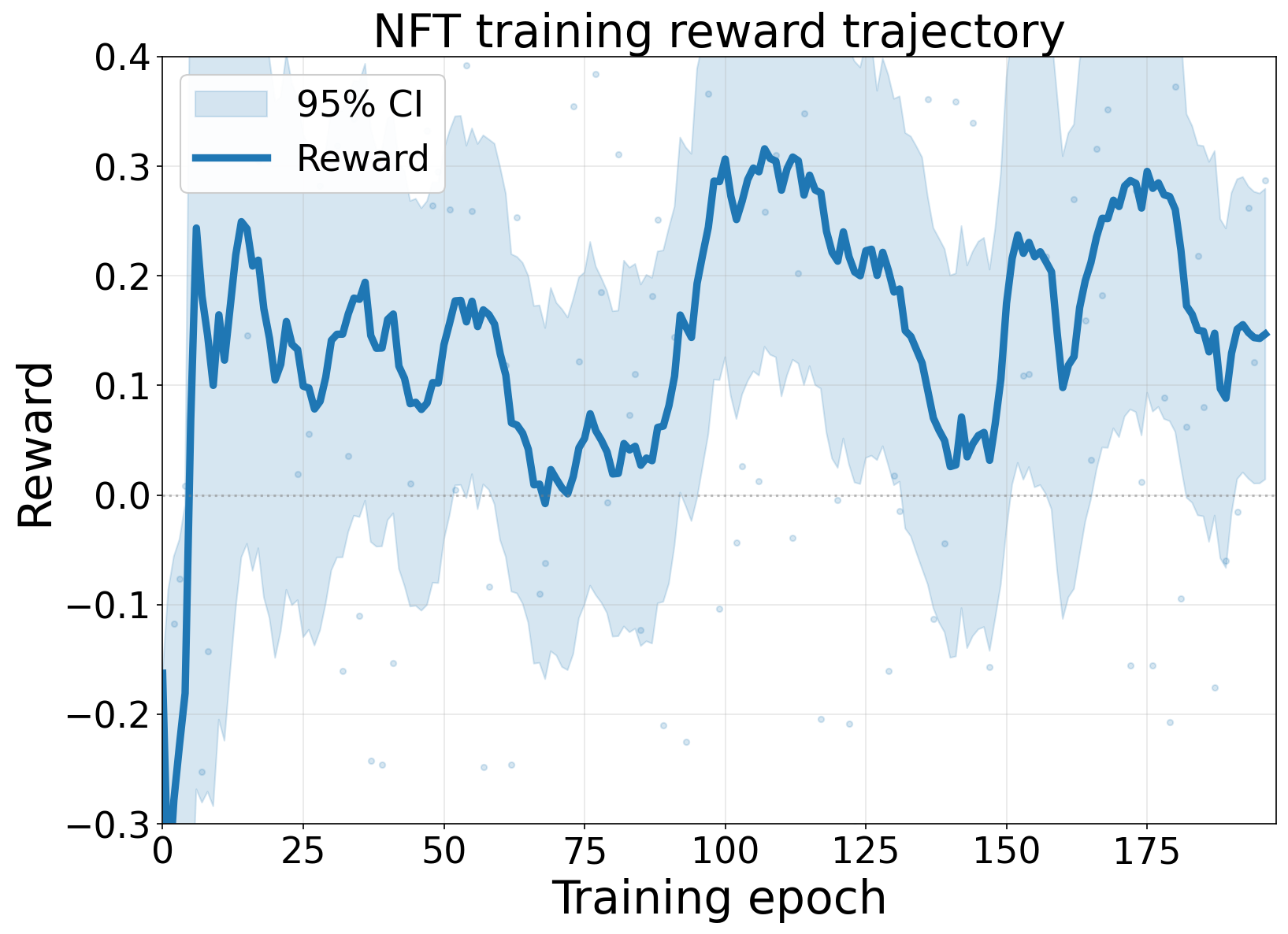}
\end{minipage}\hfill
\begin{minipage}{0.48\linewidth}
\centering
\includegraphics[width=\linewidth]{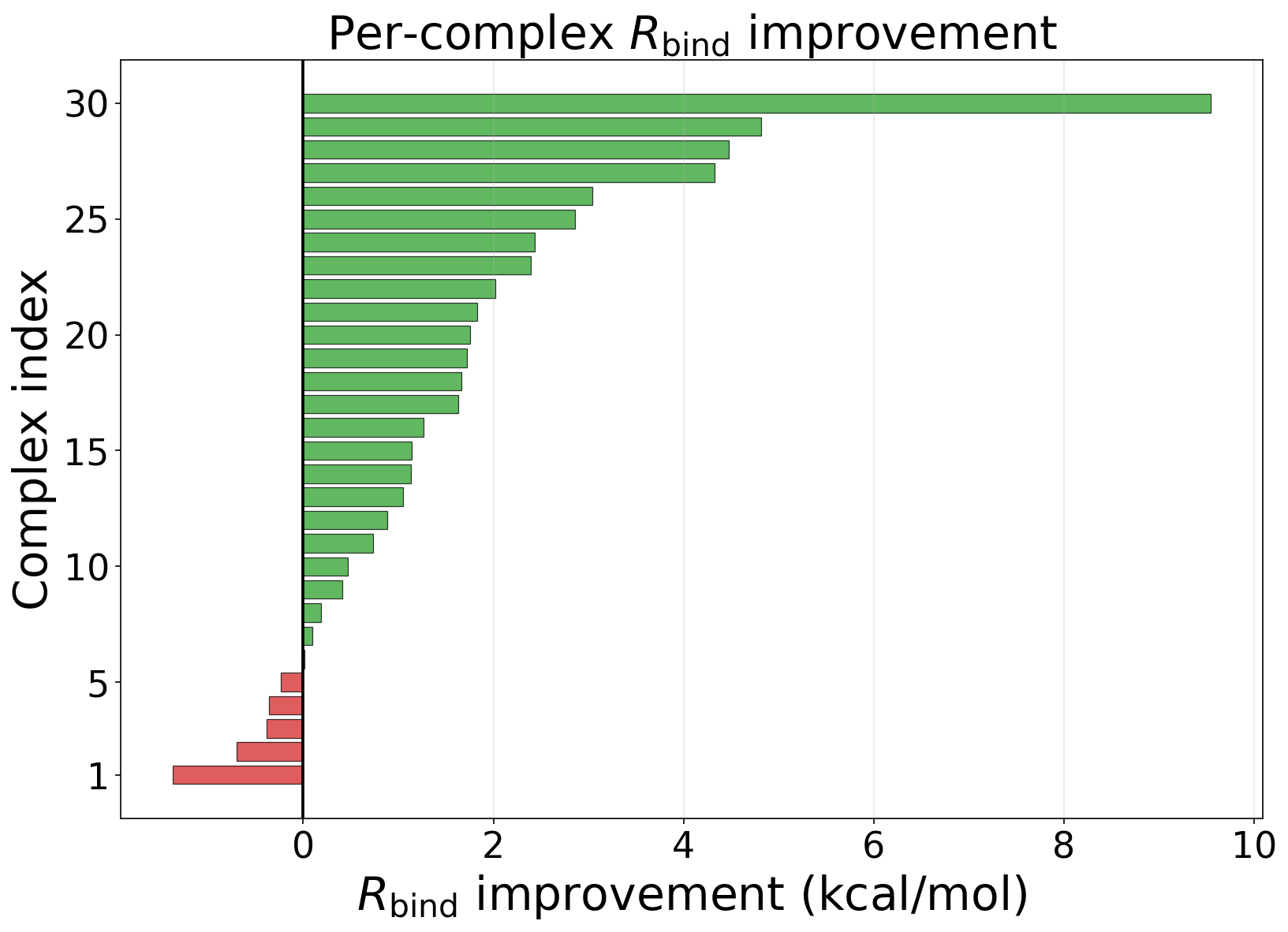}
\end{minipage}
\small
\caption{\textbf{RL training dynamics.} \textbf{Left:} mean per-batch reward over $200$ epochs (shading: $95\%$ CI). \textbf{Right:} per-complex binding-score improvement (kcal/mol; positive is better), sorted by magnitude.}
\label{fig:nft_dynamics}
\end{figure}

\paragraph{Training Dynamics.}
Figure~\ref{fig:nft_dynamics} shows how RL improves the cooperative-binding objective during training. The mean per-batch reward rises from a negative initial value to a positive plateau within approximately $25$ epochs and remains stable thereafter (left). Improvement is also observed across individual complexes: comparing each complex's first and last training visits, $R_\text{bind}$ decreases for $25$ of $30$ complexes ($83\%$), with the largest improvement reaching approximately $9.5$\,kcal/mol (right). The remaining $5$ complexes have increases smaller than $1.5$\,kcal/mol. Thus, the overall reward gain is accompanied by more favourable binding scores for most training complexes.

\raggedbottom
\setlength{\intextsep}{6pt plus 2pt minus 2pt}

\FloatBarrier

\begin{table}[H]
\centering
\caption{\textbf{Stacking-weight sensitivity and interface complementarity.} Mean geometric measures at three stacking weights $f$. Surface areas are reported per ligand; single and stacked refer to the corresponding interface calculations.}
\label{tab:stacking_sensitivity}
\small
\setlength{\tabcolsep}{4pt}
\renewcommand{\arraystretch}{1.15}
\begin{tabular*}{\linewidth}{@{\extracolsep{\fill}}cccccccc@{}}
\toprule
\textbf{Weight} & \textbf{Ligand area} & \multicolumn{2}{c}{\textbf{Buried area} (\AA$^2$)} & \multicolumn{2}{c}{$\phi_{\mathrm{bury}}$} & \multicolumn{2}{c@{}}{$S_c$} \\
\cmidrule(lr){3-4}\cmidrule(lr){5-6}\cmidrule(l){7-8}
$f$ & (\AA$^2$) & Protein & Neighbours & Single & Stacked & Single & Stacked \\
\midrule
0.2 & 688 & 340 & 230 & 0.507 & 0.836 & 0.559 & 0.662 \\
0.5 & 686 & 347 & 236 & 0.514 & 0.853 & 0.554 & 0.657 \\
0.8 & 689 & 340 & 246 & 0.507 & 0.855 & 0.540 & 0.644 \\
\bottomrule
\end{tabular*}
\end{table}

\paragraph{Stacking-Weight Sensitivity.}
We examine how the stacking weight $f$ affects surface burial and interface complementarity (Table~\ref{tab:stacking_sensitivity}). Solvent-accessible surface areas are computed with the Shrake--Rupley algorithm~\cite{shrake1973environment} using a $1.4$\,\AA\ probe. For each ligand, we separate burial by the protein from the additional burial by neighbouring ligand copies; the buried fraction $\phi_{\mathrm{bury}}$ normalises the buried area by the isolated ligand's surface area. The single-ligand reference evaluates each pose without neighbouring copies, whereas the stacked calculation includes them. The Lawrence--Colman shape complementarity $S_c$~\cite{lawrence1993shape} compares the fibril groove with either the single ligand or the exposed surface of the ligand column.

Across the three weights, including neighbouring copies increases the buried fraction from approximately $0.51$ to $0.84$--$0.86$, while the ligand column has higher groove complementarity ($S_c=0.64$--$0.66$) than the single ligand ($0.54$--$0.56$). Increasing the weight raises the area buried against neighbouring copies from $230$ to $246$\,\AA$^2$, accompanied by a modest decrease in stacked $S_c$ from $0.662$ to $0.644$. These trends reflect a trade-off between inter-ligand packing and complementarity to the fibril groove.

\subsection{Evaluation on General Protein--Ligand Docking}
\label{app:posebench}

\begin{figure}[H]
\centering
\includegraphics[width=\linewidth]{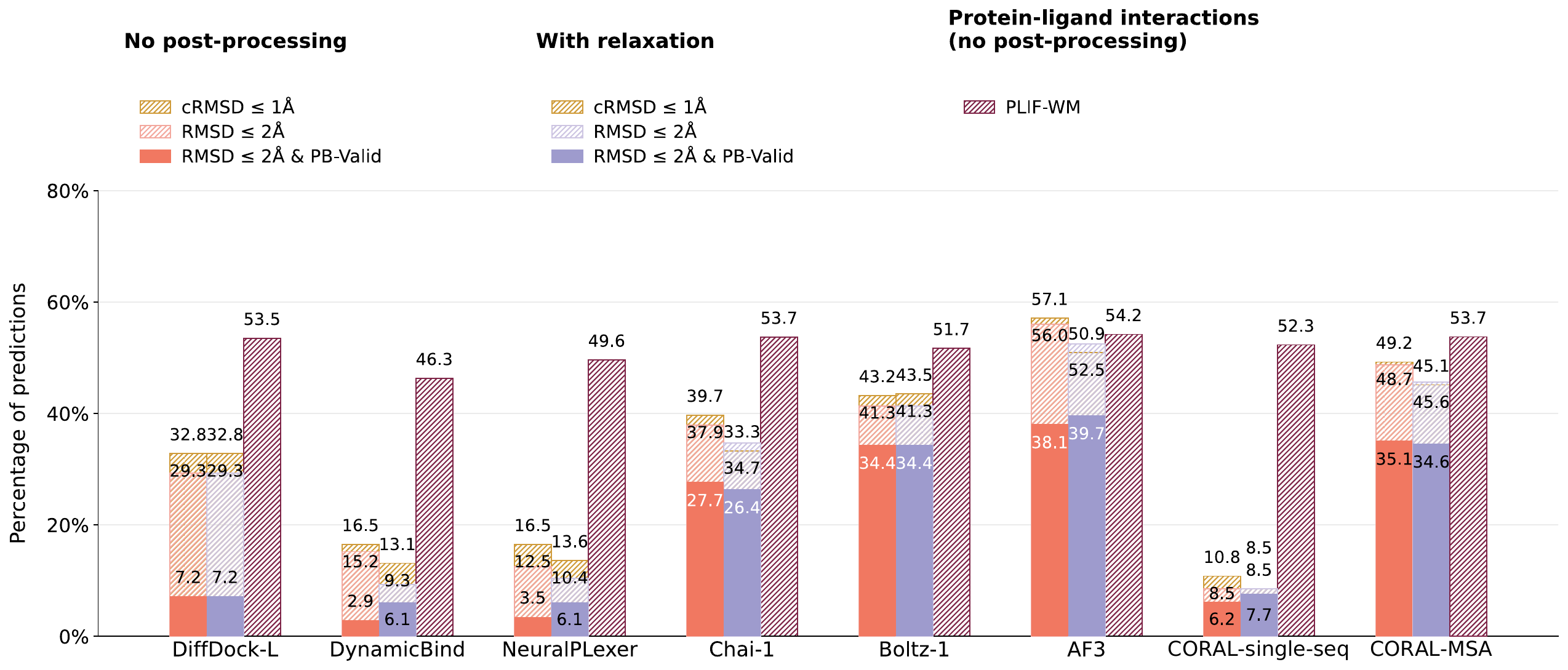}
\small
\caption{\textbf{General protein--ligand docking on PoseBusters.} Each method's three bars show pose accuracy before relaxation, after relaxation, and protein--ligand interaction similarity (PLIF-WM), respectively. Baseline results are from Figure~4 of \citet{morehead2025posebench}.}
\label{fig:posebench_comparison}
\end{figure}

To assess docking beyond amyloid fibrils, we evaluate two training and input configurations on the 130-target PoseBusters subset of \citet{morehead2025posebench} (PDB depositions after 2021-09-30): the single-sequence model described above (\textbf{CORAL-single-seq}) and \textbf{CORAL-MSA}, trained on mixed PDB and amyloid data and evaluated with multiple-sequence alignments (MSAs). Blind inference uses no reference receptor or pocket constraints and selects the highest-confidence of five samples (200 diffusion steps, 10 recycling cycles). Relaxation and interaction scoring follow the PoseBench protocol, with all 130 targets included in the averages. Without relaxation, CORAL-MSA achieves $48.7\%$ RMSD $\leq 2$\,\AA\ and $35.1\%$ RMSD $\leq 2$\,\AA\ $\wedge$ PB-Valid, compared with $8.5\%$ and $6.2\%$ for CORAL-single-seq (Figure~\ref{fig:posebench_comparison}). PLIF-WM scores are similar ($52.3$ for CORAL-single-seq and $53.7$ for CORAL-MSA). OpenMM relaxation does not further improve CORAL-MSA's pose success rates on this benchmark.

\clearpage
\section{Additional Qualitative Results}
\label{app:more_qual}

We provide additional qualitative visualisations of \modelname's predicted fibril--ligand poses on \textit{AmyDock-CryoEM} and \textit{AmyDock-Synthetic} targets, complementing the comparisons in \S\ref{sec:qualitative}.

\begin{figure}[!htbp]
\centering
\includegraphics[width=0.90\linewidth]{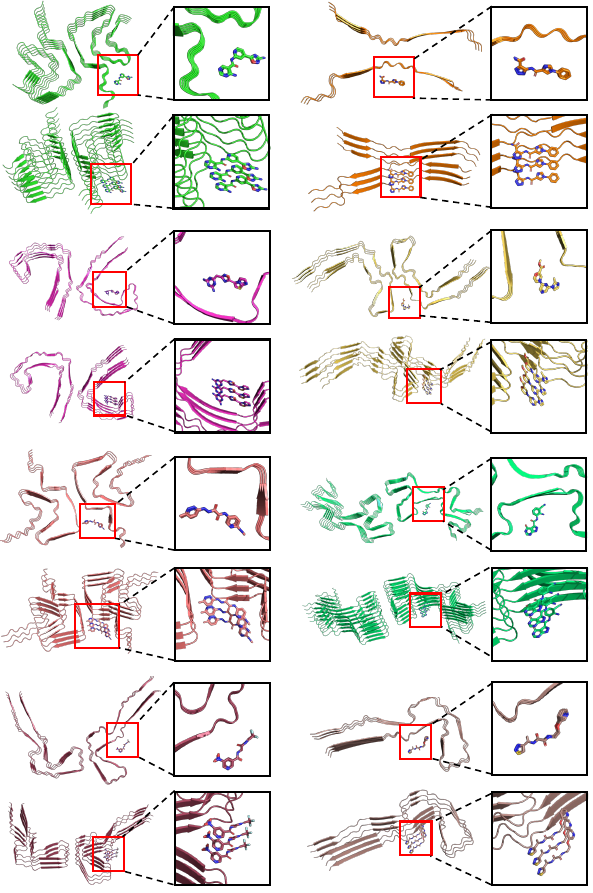}
\small
\caption{Additional qualitative visualisations of \modelname's predicted fibril--ligand poses across \textit{AmyDock} targets. Each case pairs a top-down view along the fibril axis with an oblique view, highlighting the local geometry of cooperative stacking and protein--ligand contacts.}
\label{fig:sup_add_qual_1}
\end{figure}

\begin{figure}[!htbp]
\centering
\includegraphics[width=0.90\linewidth]{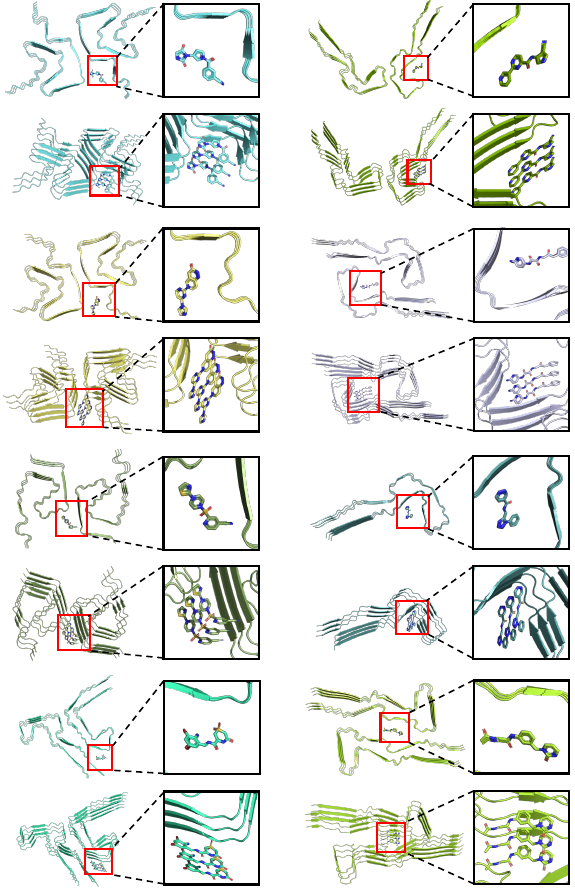}
\small
\vspace{-0.4em}
\caption{Additional qualitative visualisations of \modelname's predicted fibril--ligand poses across \textit{AmyDock} targets. Each case pairs a top-down view along the fibril axis with an oblique view, highlighting the local geometry of cooperative stacking and protein--ligand contacts.}
\label{fig:sup_add_qual_2}
\end{figure}

\begin{figure}[!htbp]
\centering
\includegraphics[width=0.90\linewidth]{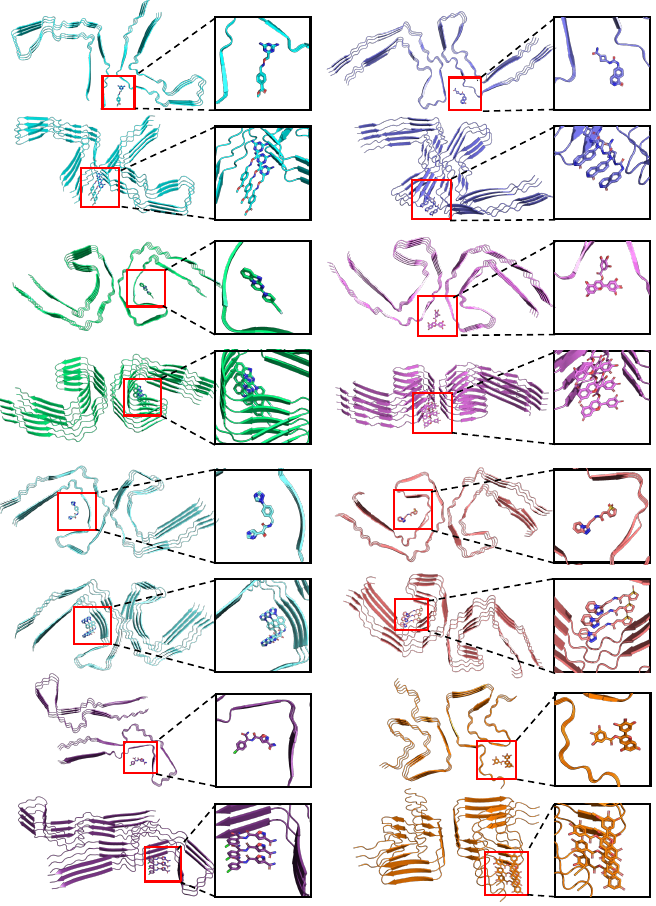}
\small
\caption{Additional qualitative visualisations of \modelname's predicted fibril--ligand poses across \textit{AmyDock} targets. Each case pairs a top-down view along the fibril axis with an oblique view, highlighting the local geometry of cooperative stacking and protein--ligand contacts.}
\label{fig:sup_add_qual_3}
\end{figure}

\FloatBarrier

\end{document}